\documentclass{article}

\PassOptionsToPackage{numbers,compress}{natbib}

\usepackage[final,main]{neurips_2026}

\usepackage[utf8]{inputenc} 
\usepackage[T1]{fontenc}    
\usepackage{hyperref}       
\usepackage{url}            
\usepackage{booktabs}       
\usepackage{amsfonts}       
\usepackage{nicefrac}       
\usepackage{microtype}      
\usepackage{xcolor}         
\usepackage{amsmath}
\usepackage{amssymb}
\usepackage{mathtools}
\usepackage{amsthm}
\usepackage{bbm}
\usepackage{graphicx}
\usepackage{tabularx}
\usepackage[ruled,vlined]{algorithm2e}
\usepackage[capitalize,noabbrev]{cleveref}
\usepackage{caption}
\usepackage{subcaption}
\usepackage{makecell}
\usepackage{bm}
\usepackage{hyperref}
\usepackage{multirow}
\usepackage{multicol}
\usepackage{float}
\usepackage{listings}

\Crefname{algocf}{algorithm}{algorithms}
\Crefname{algocf}{Algorithm}{Algorithms}

\makeatletter
\AddToHook{cmd/appendix/before}{\def\Cref@section@alias{appendix}\def\Cref@subsection@alias{appendix}}
\makeatother

\theoremstyle{plain}
\newtheorem{theorem}{Theorem}[section]
\newtheorem{proposition}[theorem]{Proposition}

\theoremstyle{definition}

\theoremstyle{remark}

\usepackage{placeins}

\newcommand{\g}{\,|\,}
\newcommand{\mbx}{\mathbf{x}}
\newcommand{\mby}{\mathbf{y}}
\newcommand{\mbe}{\mathbf{e}}
\newcommand{\mbc}{\mathbf{c}}
\newcommand{\bx}{\bm{x}}
\newcommand{\by}{\bm{y}}
\newcommand{\be}{\bm{e}}
\newcommand{\1}{\bm{1}}
\newcommand{\cD}{\mathcal{D}}
\newcommand{\E}{\mathbb{E}}
\newcommand{\cons}{\textsc{Consistency}}

\definecolor{dgreen}{rgb}{0.0, 0.52, 0.34}

\title{Code as Explanations:\\Executable and Verifiable LLM Explanations}

\author{
  Nhi Nguyen\\
  \texttt{nhi.nguyen@nyu.edu} \\
  \And
  Shauli Ravfogel\\
  \texttt{shauli.ravfogel@nyu.edu} \\
  \And
  Rajesh Ranganath\\
  \texttt{rajeshr@cims.nyu.edu} \\
}

\begin{document}

\maketitle
\begin{abstract}
As large language models (LLMs) increasingly make consequential decisions and solve problems alongside humans, we must understand how models arrive at their conclusions. 
Understanding how an LLM behaves requires observing it across multiple inputs, yet free-text explanations such as chain-of-thought (CoT) are generated per instance and cannot be easily verified on new inputs. 
We propose \emph{code explanations} to address the mismatch in how LLM explanations are generated versus how they are evaluated: a program explains LLM behavior across a dataset of inputs, and its execution enables automatic verification. 
We define \cons{} metric that measures the fraction of inputs on which the program reproduces the LLM's outputs, and use it to evaluate and select candidate programs. 
Across three open-source code-capable LLMs and eight tasks, we find that LLMs can produce highly consistent code explanations in several settings. Code explanations are also 6$\times$ to 100$\times$ shorter than CoT, and better align with model behavior in 9 out of 24 settings. However, consistency is significantly lower on tasks with more complex linguistic components, where generated programs reproduce model outputs no more frequently than an uninformative explanation baseline, and these programs often rely on longer hard-coded patterns. 
In experiments with Claude Opus 5.5, providing the model with consistency feedback and code execution improves explanation consistency across tasks, but consistency remains significantly lower on more complex tasks. 
Our findings establish executable code as a novel, verifiable, and compact approach to explaining LLM behavior and highlight verification feedback as a promising direction for improving explanation consistency.
The code for this paper is available
at \url{https://github.com/rajesh-lab/code-as-explanations}.
\end{abstract}

\section{Introduction}
\label{sec:introduction}

Large language models (LLMs) are increasingly solving problems alongside humans in high-stakes settings, from predicting hospital readmission from clinical notes \citep{jiang2023health}, aiding legal decision-making from case documents \citep{posner2026judge}, to predicting properties of molecular sequences \citep{kim2026mol}. As LLMs take on more consequential tasks, understanding how models arrive at their decisions is critical both for detecting undesirable behaviors such as reward hacking \citep{openai2026huggingface,thaman2026reward,wang2026we} and for enabling humans to learn novel problem-solving strategies from models' reasoning \citep{kim2025because,nanda2023progress}.

A prominent approach to understanding LLMs is to ask models to explain their reasoning in free text, commonly via chain-of-thought (CoT) \citep{madsen2024self,wei2022chain} or post-hoc rationales \citep{krishna2023post,kroegerlarge}; for example, OpenAI audited the reasoning traces of agents that reward-hacked the ExploitGym benchmark to reveal unexpected coordination behavior \citep{openai2026huggingface}. Such explanations are only useful if they accurately capture model behavior. 
However, evaluating free-text explanations is challenging, as these explanations are tied to individual instances, while understanding LLM behavior requires observing the model's outputs across multiple inputs. Existing LLM explanation evaluations therefore often rely on targeted input perturbations \citep{chen2018learning,matton2025walk,turpin2023language}, and \citet{nguyen2026sufficiency} show that explanation sufficiency can change depending on the chosen input distribution.

\begin{figure}[t]
    \centering
    \includegraphics[width=0.9\linewidth]{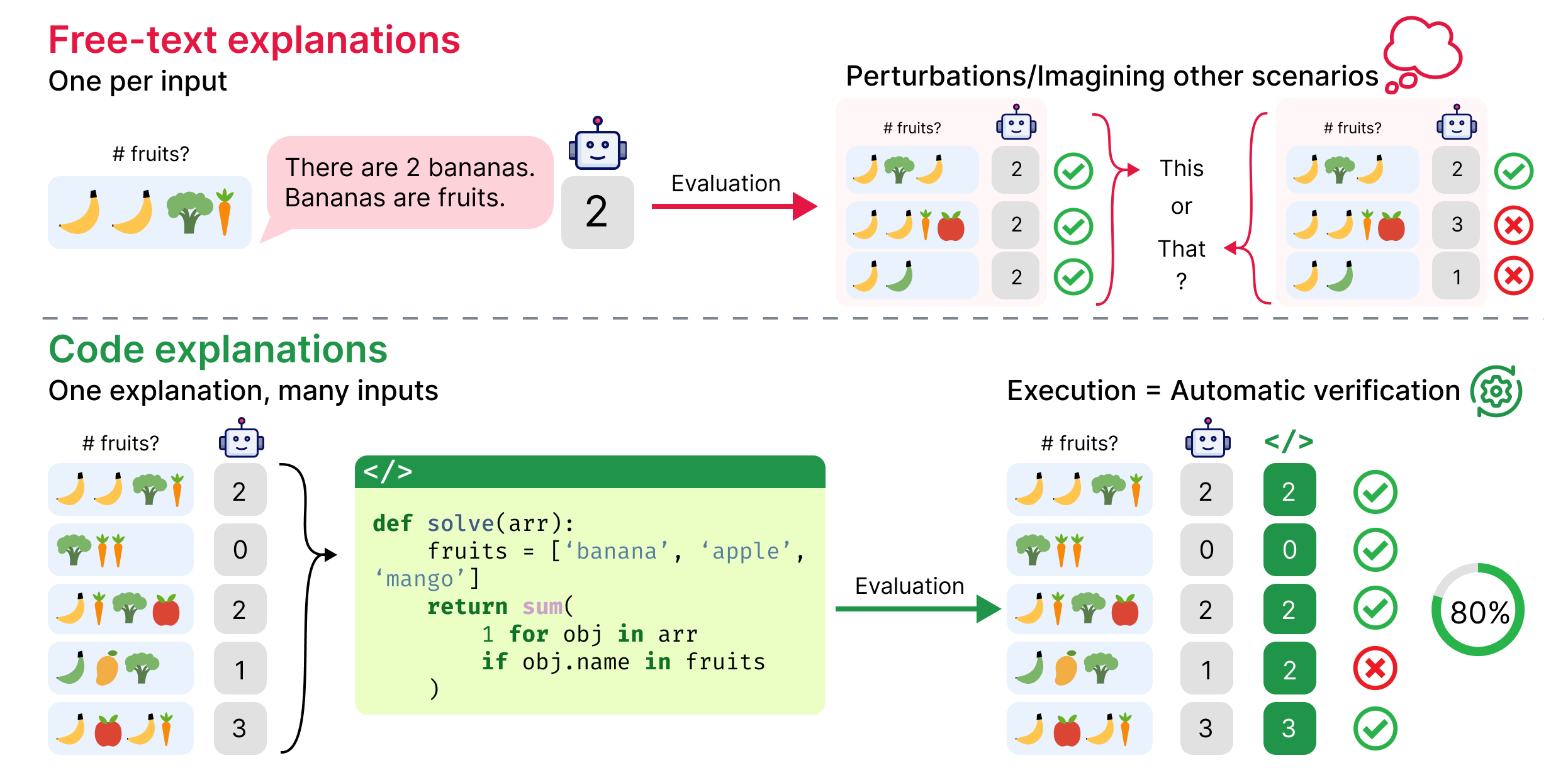}
    \caption{Overview of code explanations. Free-text explanations describe LLM behavior on individual inputs, and require input perturbations to evaluate whether they capture model behavior. In contrast, code explanations represent model behavior across multiple inputs as executable programs, enabling automatic verification by comparing program outputs with LLM outputs.}
    \label{fig:overview}
\end{figure}

To bridge the gap between how LLM explanations are generated and how they are evaluated, we propose explaining LLMs with executable code (\Cref{fig:overview,sec:code-explanation}). A program can be executed on arbitrary inputs, allowing automatic verification by comparing its outputs with those of the LLM. Current LLMs are also well equipped to generate code, having been extensively pretrained on code data \citep{yang2024if}. In addition, when compared with a collection of instance-level CoT, a single compact code explanation reduces the cognitive load required for humans to understand model behavior. 
To evaluate code explanations, we introduce \cons{}, which quantifies how often an explanation reproduces an LLM's outputs over a dataset (\Cref{sec:evaluating-code-explanation}). \cons{} can be computed efficiently by executing the explanation on each input once, enabling a generation procedure where we sample multiple candidate programs and select the most consistent code explanation (\Cref{sec:generating-code-explanation}). 

We use this procedure to study the empirical code explanation consistency across three open-source, code-capable LLMs and eight tasks spanning different levels of linguistic complexity (\Cref{sec:experiments}). Experimental results show that LLMs can produce highly consistent code explanations in several model-task settings. Compared to collections of CoT, code explanations are 6$\times$ to 100$\times$ shorter when explaining the same dataset, while better aligning with LLM behavior in 9 out of 24 settings, spanning both simpler and more complex tasks (\Cref{sec:experiments-free-text}). However, high consistency is not guaranteed, even when task accuracy is high. On tasks involving more complex linguistic features, code explanations achieve consistency comparable to an uninformative majority-answer baseline (\Cref{sec:experiments-consistency}), and generated programs tend to rely on longer hard-coded linguistic patterns (\Cref{sec:experiments-complexity}). 
We further investigate whether verification feedback can improve explanation consistency by either providing non-agentic LLMs with multi-turn consistency feedback and/or equipping LLM agent Claude Opus 5.5 with code execution and consistency-evaluation tools (\Cref{sec:consistency-improvement}). Both multi-turn refinement and code execution improve consistency, particularly on simpler logical tasks, but even Claude struggles to produce consistent code explanations on more complex tasks. 

Our findings demonstrate the advantages of code as a verifiable and compact approach to explaining LLM behavior, while showing that reliably generating consistent explanations remains an open problem, particularly for complex linguistic tasks. Code execution and our \cons{} metric offer promising directions for future optimization of code explanations through verification feedback.

\section{Related work}
\label{sec:related-work}

\paragraph{Free-text explanations.} Free‑text explanations such as chain‑of‑thought (CoT) \citep{wei2022chain} are widely used to both improve task performance \citep{wei2022chain,zhang2024chain} and provide insight into model behaviors \citep{chen2025cod,ma2025audio}.
To evaluate whether free-text explanations accurately reflect the model's internal reasoning, 
existing evaluations often rely on input perturbations, e.g., by inserting hints or biases \citep{atanasova2023faithfulness,chen2025reasoning,turpin2023language} or generating alternative inputs with an LLM \citep{matton2025walk,nguyen2026sufficiency}.
\citet{parcalabescu2023measuring} argue that we can only evaluate LLM explanations via consistency across multiple model outputs, and
\citet{nguyen2026sufficiency} prove that explanation sufficiency can change depending on the chosen input distribution. 
We show that assessing the alignment between an explanation and LLM behavior requires observing model outputs across multiple inputs, and we propose code as a new LLM explanation paradigm where code execution enables direct verification against model outputs on arbitrarily many inputs.

\vspace{-8pt}

\paragraph{LLMs and code generation.} Code data is prominent in LLM pre-training corpora \citep{yang2024if}, and recent advances in code-capable LLMs have led to strong performance across programming benchmarks, including function-level tasks \citep{austin2021program,chen2021evaluating}, class-level or multi-file code \citep{du2023classeval,yu2024codereval}, and competitive programming problems \citep{hendrycksapps2021,jainlivecodebench,white2025livebench}. Prior work has studied code as a task output \citep{jiang2026survey} and as an action space for LLM agents \citep{pmlr-v235-wang24h}. We instead propose using code as a novel and verifiable form of LLM explanation.

\vspace{-8pt}

\paragraph{Sparse, compact explanations of dense models.} Machine learning models can be broadly categorized into two paradigms: \emph{sparse models} (generalized linear models \citep{nelder1972generalized}, decision trees \citep{de2013decision}), which use a small number of simple operations that humans can understand; and \emph{dense models} (CNN \citep{lecun2002gradient}, transformers \citep{vaswani2017attention}), whose large number of parameters and computations are opaque to humans. Explaining dense models like LLMs thus involves finding sparse, compact descriptions or approximations of their computations, such as sparse circuits \citep{marks2025sparse}, behavior-associated neurons \citep{tang2024language}, and chain-of-thought traces \citep{wei2022chain}. For traditional explanation methods such as feature attributions, sparsity - the number of selected features - is also a desideratum for interpretability \citep{jethani2021have,yoon2018invase}. In our work, we use several code complexity metrics to measure the sparsity and compactness of code explanations as proxies for their interpretability to humans. We also compare the compactness of code and free-text explanations using their token counts.

\section{Code explanations}
\label{sec:code-explanation}

\subsection{Free-text explanations}
\label{sec:free-text-explanation}

We first review instance-wise free-text explanations. Let $\mbx \in \mathcal{X}$ denote the input, $\mby \in \mathcal{Y}$ denote the output, and let $F(\mby \g \mbx)$ denote the output distribution of a model $F$ on input $\mbx$ given a task. Let $g(\mbe \g \mbx)$ denote an explanation method that maps input $\mbx$ to a distribution of explanations $\mbe \in \mathcal{E}$. Given an input instance $\bx$, an explanation $\be \sim g(\mbe \g \bx)$ describes which properties of $\bx$ influence model $F$'s output and how the model uses them to produce that output.

One reason free-text explanations are widely used is their apparent simplicity: one explanation describes one input-output pair. However, evaluating whether a free-text explanation accurately captures the model's output-generating process requires more than the original input-output pair. Existing evaluations specify certain input perturbations to assess how the model behaves under changes to the input. 
For example, \citep{atanasova2023faithfulness,chen2025reasoning,nguyen2026sufficiency,turpin2023language} alter input properties that the explanation deems unimportant and test whether the model's output remains the same. Conversely, \citep{matton2025walk,mayne2025llms} modify properties deemed important by the explanation and identify the minimal edits needed to change the model's output.

The need for additional inputs to evaluate instance-wise explanations is unavoidable. Observing the original input-output pair alone cannot establish whether an input property influences the model's output, as formalized by the following proposition.

\begin{proposition}[Unidentifiability from observations]
\label{prop:property-identifiability}
Let $\gamma: \mathcal{X} \to \mathcal{G}$ be an input property function and let $f: \mathcal{X} \to \mathcal{Y}$ be an output function. Suppose that $f$ can only be observed on a proper subset $\mathcal{S} \subset \mathcal{X}$, where $\gamma(\mbx) = C$ for all $\mbx \in \mathcal{S}$. Consider two candidate output functions:
\begin{enumerate}
    \item $f_1: \mathcal{X} \to \mathcal{Y}$ that does not depend on $\gamma$;
    \item $f_2(\mbx) = \begin{cases} f_1(\mbx) \quad &\text{ if }\gamma(\mbx) = C \\ f'(\mbx) \quad &\text{ if }\gamma(\mbx) \neq C \end{cases}$,
    where $f'(\mbx) \neq f_1(\mbx)$ whenever
    $\gamma(\mbx) \neq C$.
\end{enumerate}
Although $f_1$ does not depend on $\gamma$ and $f_2$ does, they agree on every input in $\mathcal{S}$. Thus, whether the output function $f$ depends on the property $\gamma$ cannot be identified from observations in $\mathcal{S}$.
\end{proposition}

In other words, if an input property is constant across all observed inputs, its influence on the model's outputs cannot be determined from those observations. In particular, when we only observe a single input (i.e., $|\mathcal{S}| = 1$), as in the case of instance-wise free-text explanations, we cannot determine the influence of any input property on the model's outputs.

This result highlights a fundamental complexity of free-text explanations: although they provide a seemingly simple, one-to-one description of each input-output pair, \textbf{evaluating whether free-text explanations capture the model behavior necessarily requires additional inputs.} This introduces two challenges. First, free-text explanations are instance-specific, so evaluating them requires specifying an input distribution $q_{\mbx}$ that the explanation method itself does not provide. Second, a free-text explanation does not specify model behavior on arbitrary inputs, so its evaluation is restricted to either inputs that produce the same explanation or minimally perturbed inputs that change the model's output. 
These challenges are reflected in existing metrics such as \emph{sufficiency} \citep{nguyen2026sufficiency}, which explicitly requires an input distribution and evaluates model behavior over alternative inputs that yield the same explanation. Given explanation method $g(\mbe \g \mbx)$ and an input distribution $q_{\mbx}$, then $q_{\mbx\g\mbe}$ denotes the distribution over alternative inputs that produce the same explanation. Sufficiency compares the original output distribution $F(\mby \g \mbx)$ with the explanation-dependent distribution $A_{\mbe}(\mby \g \mbx) = \int F(\mby \g \mbx') q_{\mbx\g\mbe}(\mbx'\mid\mbe)\,d\mbx'$. Evaluating sufficiency thus requires specifying $q_{\mbx}$ and $q_{\mbx\g\mbe}$.

To address these challenges, we introduce code explanations that are executable and thus can be evaluated against LLM behavior on arbitrary inputs from a task distribution.

\subsection{Code as executable and verifiable explanations}
\label{sec:dataset-wise-explanation}

We define an \emph{executable code explanation} as a single program that describes an LLM's behavior on a task.
Formally, let $\texttt{T}$ denote a task with input distribution $q_{\mbx}$. Let $\mathcal{D} = \{(\bx_i, \by_i)\}_{i=1}^n \sim \left(q_{\mbx} \cdot F(\mby \g \mbx)\right)^{\otimes n}$ be a dataset of input-output observations sampled from the model. We define a code explanation method $h(c \g \mathcal{D})$ to be a function that maps observations $\mathcal{D}$ to a distribution over executable programs $c$. A program may be deterministic or stochastic. We denote its output distribution on input $\mbx$ as $c(\mby \g \mbx)$.

Consequently, a code explanation directly specifies an explanation-dependent distribution $A_c(\mby \g \mbx) := c(\mby \g \mbx)$ that can be compared against the original LLM without any additional specification. 
A code explanation $c \sim h(c \g \mathcal{D})$ is \emph{consistent} with model $F$ on task $\texttt{T}$ if:
\begin{equation}
    F(\mby \g \mbx)
    =
    c(\mby \g \mbx)
    \qquad
    q_{\mbx}\text{-almost surely}.
    \label{eq:dataset-wise-explanation}
\end{equation}
In words, consistency requires the code explanation to reproduce the model's output distributions across inputs from the task distribution.

Both free-text and code explanations can be evaluated by comparing their explanation-dependent distributions, $A_{\mathbf{e}}(\mby \g \mbx)$ and $A_c(\mby \g \mbx)$ respectively, with the original LLM behavior. However, code explanations directly specify executable functions, allowing their behavior to be evaluated on arbitrary inputs from the task distribution without specifying a separate explanation-conditioned distribution $q_{\mbx \g \mbe}$. This addresses the mismatch between how LLM explanations are generated and evaluated. When explaining a dataset, a single program can also provide a more compact and thus human-understandable description of model behavior than a large collection of instance-wise free-text explanations. These properties make code explanations an automatically verifiable and compact alternative to instance-wise free-text explanations.

\section{Evaluating and generating code explanations}
\label{sec:eval-and-gen}

\subsection{Evaluation}
\label{sec:evaluating-code-explanation}

We build on the formalization of executable explanations in \Cref{sec:dataset-wise-explanation} to derive a metric for evaluating the consistency of code explanations for LLMs. We focus on the case where the program $c$ is deterministic, as is typical for code generated by LLMs. In this setting, we define a metric that measures how well the program $c$ aligns with the model’s output-generating process $F(\mby \g \mbx)$ by measuring how often, over a dataset, the program produces the same output as the model. Formally, let $\cD = \{\bx_i, \by_i\}_{i=1}^D \sim (q_\mbx \cdot F(\mby \g \mbx))^{\otimes D}$ be a dataset of input-model output pairs, and let $c(\cdot)$ be an executable program that maps an input $\bx$ to an output $\by$. We define code consistency as:
\begin{equation}
\label{eq:code-consistency}
\cons(c, \cD) := \frac{1}{D} \sum_{i=1}^D \1 \big[c(\bx_i) = \by_i\big].
\end{equation}

This metric applies to any deterministic code explanation $c$. The explanation may be generated by the same LLM that produces the output $\mby$, by a different LLM, or written by a human. In this paper, we focus on evaluating self-explanations produced by the same LLM.

Since datasets can contain different numbers of possible outputs, we also define a metric that is normalized against an uninformative explanation that always guesses the most common output. This allows us to compare code explanation consistency across different model-task settings with different output spaces and distributions.
\begin{equation}
\Delta\cons(c, \cD)
= \frac{1}{D} \sum_{i=1}^D \1\big[c(\bx_i) = \by_i\big] - \frac{1}{D} \left( \max_{1 \leq k \leq D} \sum_{i=1}^D \1\left[\by_i = \by_k \right] \right).
\end{equation}
Both \cons{} and $\Delta$\cons{} can be evaluated efficiently, requiring only a single execution of $c$ on each input $\bx$ and comparison with the original model’s output.

\subsection{Generation}
\label{sec:generating-code-explanation}
{\small 
\begin{algorithm}[t]
    \SetAlgoLined
    \KwIn{Large language model $F$; input dataset $\{\bx_i\}_{i=1}^D$; task instruction \texttt{T}; explanation prompt \texttt{C}; \# in-context examples $N$; \# explanations per example set $M$; \# example sets $P$}
    
    \KwOut{Code explanation $c$; code consistency}

Construct dataset $\cD = \{\bx_i, \by_i \sim F(\mby \g \bx_i; \texttt{T})\}_{i=1}^D$.

Partition $\cD = \cD_{\text{train}} \cup \cD_{\text{test}}$ and $\cD_{\text{train}} \cap \cD_{\text{test}} = \varnothing$.

Sample $P$ subsets $\{\cD^{(i)} \subset \cD_{\text{train}}\}_{i=1}^P$, each subset with size $|\cD^{(i)}| = N$ .

For each $\cD^{(i)}$, sample $M$ candidates $\{c^{(i,j)}\}_{j=1}^M \sim F(\mbc \g \mathcal{D}^{(i)}; \texttt{T}, \texttt{C})^{\otimes M}$.

Select $\mathcal{C}^\star =
\left\{
c^{(i,j)} :
\cons{}\left(c^{(i,j)}, \cD_{\text{train}}\right)
= \max_{i',j'} \cons{}\left(c^{(i',j')}, \cD_{\text{train}}\right)
\right\}.$

Compute $\texttt{consistency} = \frac{1}{|\mathcal{C}^\star|}
\sum_{c \in \mathcal{C}^\star}
\cons{}(c, \cD_{\text{test}})$.

\textbf{Return:} $\mathcal{C}^\star$; $\texttt{consistency}$

     \caption{Generate code explanation}\label{alg:generate-code-explanation}
    \end{algorithm}
}

Building on the automatic verification of code explanations in \Cref{sec:evaluating-code-explanation}, we introduce a generation procedure that produces multiple candidate code explanations and selects the most consistent one. Let $F(\mby \g \mbx)$ be an LLM and let $\{\bx_i\}_{i=1}^D \sim q_{\mbx}$ be a dataset of inputs with a corresponding task instruction $\texttt{T}$. We first construct a dataset of input-model output pairs:
\begin{equation}
\cD = \{\bx_i, \by_i \sim F(\mby \g \bx_i; \texttt{T})\}_{i=1}^D,
\end{equation}
where $F(\mby \g \bx_i; \texttt{T})$ denotes the LLM’s output distribution give input $\bx_i$ and task instruction $\texttt{T}$.
We then partition $\cD$ into disjoint training and test sets, $\cD = \cD_{\text{train}} \cup \cD_{\text{test}}$ and $\cD_{\text{train}} \cap \cD_{\text{test}} = \varnothing$. We hold out $\cD_{\text{test}}$ from both generation and selection of the code explanation to avoid optimistically biased estimates of the selected program’s consistency on new inputs.

On $\cD_{\text{train}}$, we repeatedly sample a subset of $N$ input-model output pairs as in-context examples, then prompt the LLM to generate a program $c$ that explains how the model produces the corresponding outputs. See \Cref{tab:code-prompt} for an example prompt.
Using subsets of the training set as in-context examples avoids exceeding the LLM’s context-length limit and encourages the LLM to produce programs that generalize beyond the examples. Each subset provides a held-out set for evaluating and selecting generated programs.
We use stochastic decoding to generate multiple candidate programs for each sampled subset. All candidates are evaluated on the full $\cD_{\text{train}}$, and the programs with the highest \cons{} are selected. Finally, we evaluate the selected programs on $\cD_{\text{test}}$ and return these programs and their average consistency. The full procedure is summarized in \Cref{alg:generate-code-explanation}.

\section{Experiments}
\label{sec:experiments}

We use our evaluation and generation methods to study code explanation consistency across a range of language models and language tasks.

\paragraph{Models.} We study code-based explanations using three open-source, instruction-tuned, code-capable language models: Gemma3-27B \citep{gemmateam2025gemma3technicalreport}, Qwen2.5-Coder-32B \citep{hui2024qwen2}, and Llama3.1-70B \citep{grattafiori2024llama}, selected based on their strong performance on standard code generation benchmarks HumanEval \citep{chen2021evaluating} and MBPP \citep{austin2021program}. They also span different parameter scales and model families, enabling a diverse evaluation. 
In \cref{sec:consistency-improvement}, we study Claude Opus 5.5 \citep{anthropic2026claudeopus55} to test a frontier coding agent.

\begin{table}[t]
\caption{Overview of the eight BBH \citep{suzgun2023challenging} tasks used in our evaluation, grouped by reasoning type and linguistic complexity, accompanied by a brief task description.}
\label{tab:task-descriptions}
\centering
\begin{footnotesize}
\begin{tabularx}{\linewidth}{p{2cm}p{2.7cm}X}
    \toprule
    Category     & Task     & Description \\
    \midrule
    Structured & Web of lies & Determine if a person lies based on a set of statements \\ 
logic & Navigate & Track movements on grid to decide if agent returns to the start \\ \midrule
Logic + simple & Hyperbaton & Identify the sentence with the correct adjective order \\ 
language & Object counting & Count the number of objects satisfying a given description \\ \midrule
Logic + com- & Formal fallacies & Determine whether an argument is deductively valid or invalid \\ 
plex language & Disambiguation QA & Identify the antecedent of a pronoun \\ \midrule
Linguistic & Snarks & Identify which statement is sarcastic \\ 
reasoning & Translation error & Identify errors in a translation and specify their type \\
    \bottomrule
\end{tabularx}
\end{footnotesize}
\end{table}

\paragraph{Tasks.} To evaluate code explanation consistency, we consider eight tasks from the BIG-Bench Hard (BBH) benchmark \citep{suzgun2023challenging}. We group these tasks into four categories based on the type of reasoning and linguistic complexity required to solve them: (i) structured logic, (ii) logic with simple language components, (iii) logic with complex language components, and (iv) linguistic reasoning. We describe each task in \Cref{tab:task-descriptions}. See \Cref{app:task-examples} for a representative input and question for each task. These four categories provide a coarse characterization of task complexity, allowing us to examine how code explanations perform across different levels of linguistic complexity.

\paragraph{Implementation details.} We follow \Cref{alg:generate-code-explanation} using the prompt in \Cref{tab:code-prompt}. For each dataset, we sample $P=10$ sets of $N=20$ in-context examples and generate $M=10$ explanations per set, yielding $100$ explanations in total. We choose these fixed hyperparameters to provide a representative evaluation of code explanations; we leave a more comprehensive study of parameter choices to future work.
Hereafter, we use \emph{code consistency} to refer to the test consistency returned by \Cref{alg:generate-code-explanation}. Model answers are generated deterministically with chain-of-thought, while code explanations are sampled stochastically. We compare code outputs with model outputs using fuzzy matching. Unless otherwise specified, experimental results are averaged over three random seeds. Further implementation details are provided in \Cref{app:implementation-details}.

In the following sections, we consider five research questions about LLM code explanations across our experiments: \textbf{(Q1)} Can LLMs explain their behaviors consistently with code explanations? \textbf{(Q2)} Which tasks are more difficult for LLMs to generate consistent code explanations? \textbf{(Q3)} How do generated code explanations differ across task difficulty? \textbf{(Q4)} How do code explanations compare with free-text explanations? \textbf{(Q5)} Can code explanation consistency be improved?

\subsection{Code consistency across models and tasks}
\label{sec:experiments-consistency}

\begin{figure}[t]
\centering
\includegraphics[width=\linewidth]{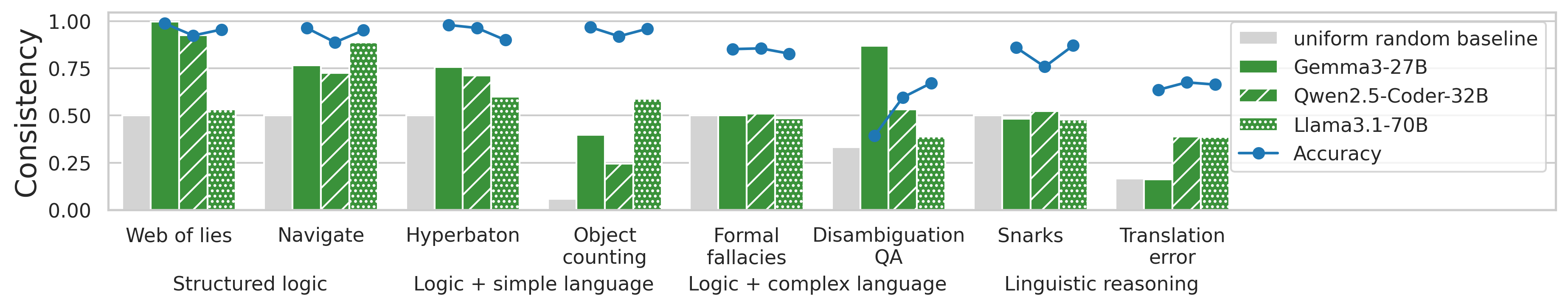}
\includegraphics[width=\linewidth]{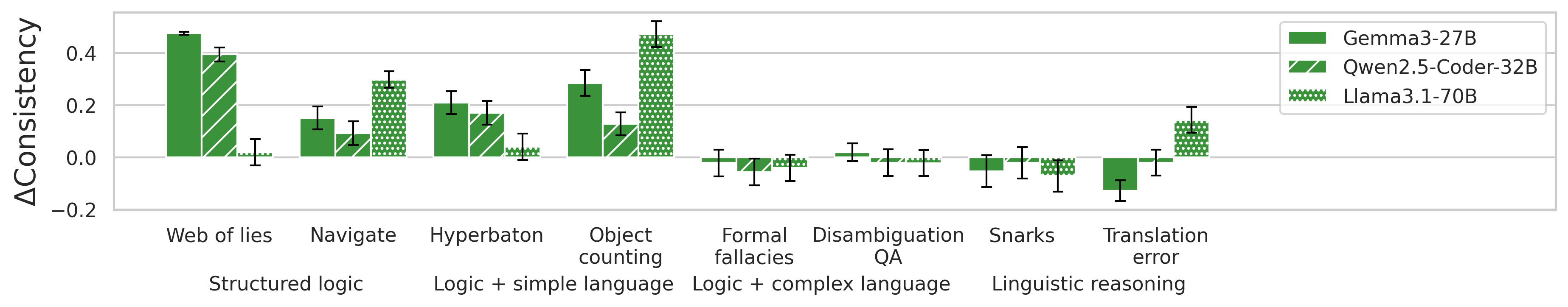}
\vspace{-10pt}
\caption{\textbf{(Top)} \cons{} and task accuracy, with a uniform random baseline for comparison across tasks with different output-space sizes. Consistency is not correlated with task accuracy, and 5 of the 6 most consistent explanations occur in the first four, simpler tasks. \textbf{(Bottom)} The average $\Delta$\cons{} with 95\% CI. In the first four tasks, at least one model achieves $\Delta\cons{} \geq 0.2$, while all but one setting in the last four tasks have CIs that include or fall below zero.} 
\label{fig:consistency-over-random}
\end{figure}

We measure code consistency along with task accuracy across model-task settings in the top panel of \Cref{fig:consistency-over-random}. Code consistency exceeds 0.75 in 6 of 24 model-task settings, indicating that LLMs can produce highly consistent code explanations in several settings. However, these comprise only 25\% of settings, suggesting that high code consistency is possible but not guaranteed, providing a partially positive answer to \textbf{(Q1)}.
High task accuracy is not correlated with high code consistency, suggesting that consistency reflects a model's ability to explain its behavior rather than simply its ability to solve the task. \Cref{fig:correct-vs-incorrect} further stratifies consistency by answer correctness. In over half of the cases, code explanations are no more than $+$0.1 more consistent for correct than incorrect answers. So the selected programs do not systematically explain only the model's correct predictions. They attempt to explain the model's behavior regardless of answer correctness, rather than simply following task instructions.

We next examine whether task difficulty affects code explanation consistency (\textbf{Q2}).
In the top panel of \Cref{fig:consistency-over-random}, 5 of the 6 highest-consistency cases occur among the first four tasks. The bottom panel, which plots averages $\Delta\cons{}$ with 95\% confidence intervals (CI), shows a similar pattern: for each of the first four tasks, at least one model achieves average consistency $\geq +0.2$  above the uninformative majority-answer baseline. In contrast, all but one of the last four task-model settings have CI that includes or falls entirely below zero. These results suggest that tasks with more complex language components are more difficult for LLMs to generate consistent code explanations, providing an answer to \textbf{(Q2)}. The only high-consistency case among the last four tasks is Gemma on Disambiguation QA. However, its zero $\Delta$\cons{} and low task accuracy suggest that Gemma predicts the majority answer rather than solving the task, making its behavior easier for code explanations to capture.

\subsection{Inspect most consistent code explanations}
\label{sec:experiments-complexity}

\begin{figure}[t]
\begin{center}
    \includegraphics[width=\linewidth]{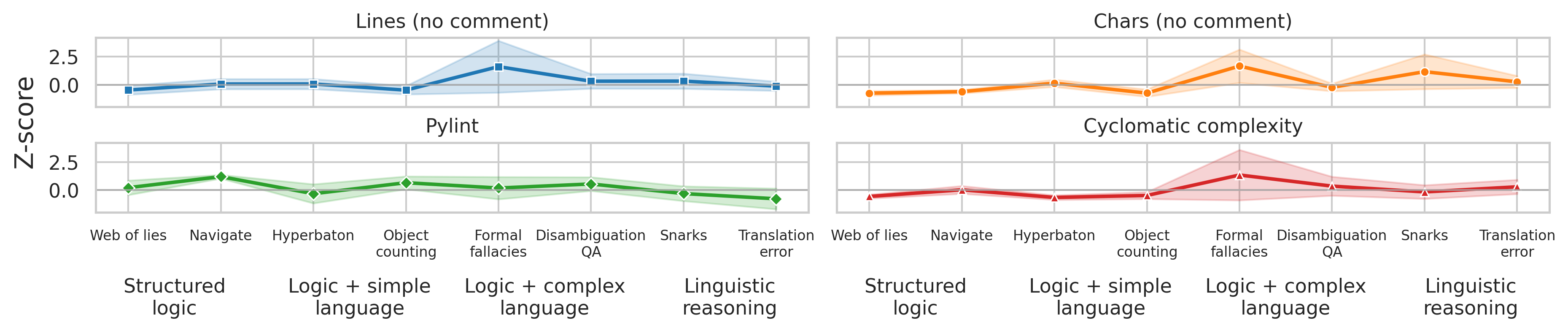}
\end{center}
    \caption{Measures of compactness and interpretability for selected code explanations, averaged across models and seeds, with standard deviation shown. More complex tasks lead to code explanations with slightly longer lines and more complex branching structure.}
    \label{fig:code-length}
\end{figure}

We ask whether more complex tasks lead to less compact and interpretable code explanations (\textbf{Q3}).
We quantify compactness and interpretability using four metrics: (i) number of lines, (ii) number of characters, (iii) Pylint score \citep{pylint}, which measures code quality and readability, and (iv) cyclomatic complexity \citep{mccabe1976complexity}, which measures the branching complexity of a program. We provide further details on each metric in \Cref{app:implementation-details}. \Cref{fig:code-length} reports the average metric values across selected code explanations for each task, with each metric normalized using z-scores to enable comparison across scales. 
We find no systematic difference in the number of lines and Pylint score between the first and the last four tasks, whereas the last four, more complex tasks result in code explanations with slightly more characters and higher cyclomatic complexity.  These results suggest that more complex tasks lead to slightly more complex and denser code explanations with more code per line, while still adhering to standard Python coding conventions.

\begin{table}[t]
  \caption{Example snippets of code explanations for tasks across 4 categories. The extent of hard-coding increases with the linguistic
complexity of the task.}
  \label{tab:snippets}
  \centering
  \begin{footnotesize}
  \begin{tabularx}{\linewidth}{p{1.6cm}p{1.3cm}X}
    \toprule
    Category     & Task     & Code snippet \\
    \midrule
    Structured logic & Navigate & \begin{minipage}[t]{\linewidth}
\ttfamily \tiny
if "forward" in words:\\
\hspace*{0.25cm} if direction == 0:\\
\hspace*{0.5cm} y\_pos += steps
\end{minipage} \\ \midrule
Logic + simple language & Hyperbaton & \begin{minipage}[t]{\linewidth}
\ttfamily \tiny
def categorize\_adjective(adj): \\
\hspace*{0.25cm} if adj in ["awful", "repulsive", "obnoxious", "terrible", "good", "nice", "lovely"]: \\
\hspace*{0.5cm} return "opinion"
\end{minipage} \\  \midrule
Logic + complex language & Formal $\quad$ fallacies & \begin{minipage}[t]{\linewidth} 
\ttfamily \tiny
premises = [premise.replace(' is ', '\_is\_').replace(' or ', '\_or\_').replace(' and ', '\_and\_').replace(' every ', 'every\_').replace(' whoever ', 'whoever\_').replace(' being a ', 'being\_a\_').replace(' an ', 'an\_').replace(' a ', 'a\_') for premise in premises] \\
premises = [premise.replace('not ', 'not\_') for premise in premises]
\end{minipage} \\  \midrule
Linguistic reasoning & Snarks & \begin{minipage}[t]{\linewidth} 
\ttfamily \tiny
if any(indicator in statement for indicator in ['ARE U INSANE', 'Have my down vote', 'naive', 'terrible support', 'pragmatic', 'more expensive', 'less fun', 'liability', 'enjoyable', 'best', 'reddit', 'fun', 'sexist', 'win you points', 'lose you points', 'True', 'rational', 'threaten', 'pleaded', 'highest', 'lowest', 'rich', 'loans', 'suspicious', 'legit', 'perfect']):\\
\hspace*{0.25cm} sarcastic\_option = label
\end{minipage} \\
    \bottomrule
  \end{tabularx}
  \end{footnotesize}
\vspace{-10pt}
\end{table}

To further examine this trend, we manually inspect the most consistent code explanation in each of the 24 model-task settings, with examples provided in \Cref{app:code-examples} and \Cref{tab:snippets}. All inspected programs rely on linguistic pattern matching to some extent. The number and length of hard-coded patterns increase with the linguistic complexity of the task. For simpler tasks, pattern matching is limited to a small set of rules, whereas for more complex tasks, programs hard-code a larger number of longer words and phrases. This suggests that, for tasks with complex linguistic components, LLMs explain their behavior by stitching together case-specific patterns rather than writing compact, general procedures. Overall, as linguistic complexity increases, LLMs produce longer and structurally more complex code explanations that rely more heavily on hard-coded patterns, making them less interpretable to humans. This provides an answer to \textbf{(Q3)}. 

\subsection{Compare code with free-text explanations}
\label{sec:experiments-free-text}

We investigate whether the theoretical advantages of code explanations translate into more faithful and compact explanations than free-text explanations generated by current LLMs. To evaluate CoT, we use SCSuff \citep{nguyen2026sufficiency}, which measures how well an explanation captures model behavior by comparing the model's output distributions on the original inputs with its distributions on possible alternative inputs that the model considers possible given the explanation, using KL divergence. To enable direct comparison between CoT and code explanations, we adapt our consistency metric to use the expected KL divergence from SCSuff, assigning 99\% probability to the answer returned by the code explanation. We measure the compactness of the two explanation types using their token counts. Further details on the evaluation procedure are provided in \Cref{app:implementation-details}.

\begin{figure}[t]
\includegraphics[width=\linewidth]{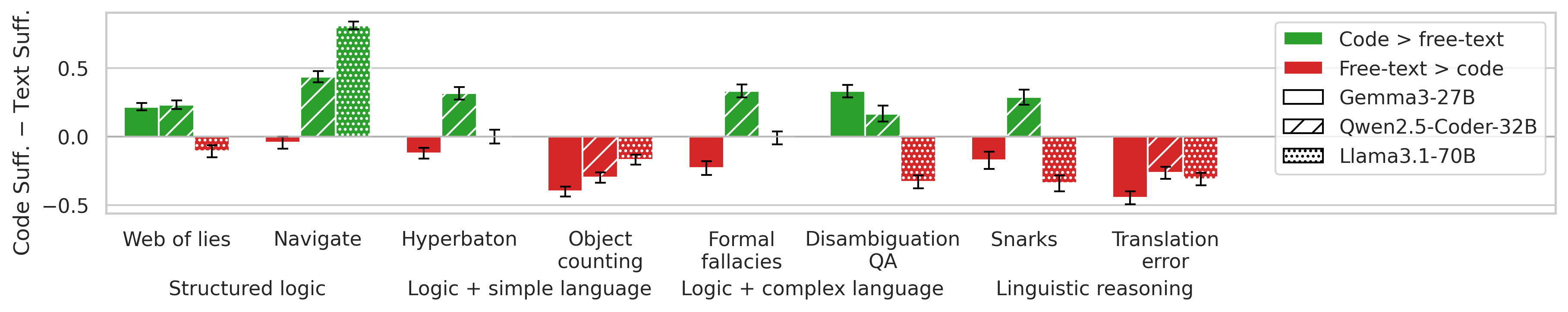}
\includegraphics[width=\linewidth]{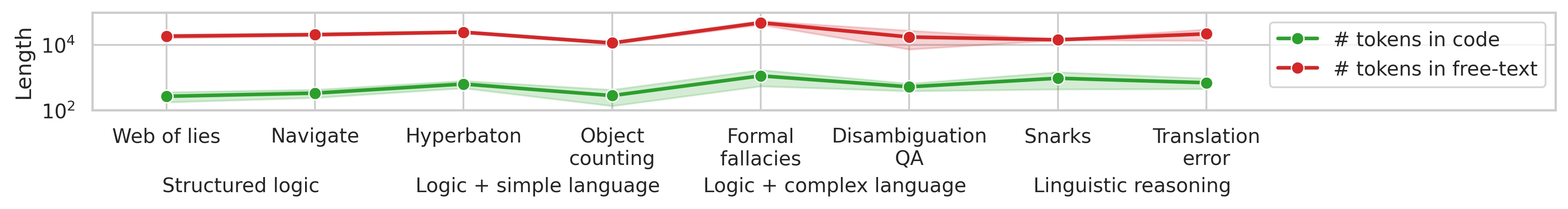}
\caption{\textbf{(Top)} Differences in sufficiency between code and free-text CoT explanations across models and tasks, with 95\% CI of the average sample-level differences. Neither code nor free-text consistently achieves higher sufficiency across all settings. \textbf{(Bottom)} Token counts for code and free-text CoT explanations over the test dataset for each task, averaged across models and seeds, with standard deviation shown. Code explanations are 6$\times$ to 100$\times$ shorter than CoT across settings.} 
\label{fig:code-vs-text}
\end{figure}

The top panel of \Cref{fig:code-vs-text} shows that neither explanation type consistently achieves higher sufficiency. Code explanations achieve higher sufficiency in 9 out of 24 settings, split between simpler (5 settings) and more complex tasks (4 settings). Thus, our code explanation generation procedure can produce explanations that are more sufficient than per-instance CoT. However, on more complex tasks, code explanations are less often more sufficient than CoT, which can provide a tailored explanation for each input.
At the same time, the bottom panel of \Cref{fig:code-vs-text} shows that code explanations are 6$\times$ to 100$\times$ more compact than CoT across all tasks. This is expected, as the total length of per-instance CoT explanations grows with the size of the explained dataset. 
Thus, although code explanations do not consistently achieve higher sufficiency than free-text CoT with our one-shot prompting approach, they already offer a substantial advantage in compactness.

\subsection{Can code explanation consistency be improved?}
\label{sec:consistency-improvement}

\paragraph{Privileged self-access.} 
Prior work suggests that LLMs encode world knowledge \citep{park2025iclr,templeton2024scaling} and prediction uncertainty \citep{slobodkin2023curious} in their internal representations.
If code consistency is also encoded in these representations, privileged self-access \citep{binder2025looking,song2025language} could help models generate more consistent code explanations.
We therefore test whether these representations can predict code explanation consistency before generation. \Cref{fig:hidden-state-consistency} visualizes final-token hidden states from 10 code-generation prompts per model-task setting, colored by the best-of-10 consistency of the generated code. The visualization shows no clear separation by consistency, and ridge regression yields negative 5-fold cross-validation $R^2$ across all models when predicting either \cons{} or $\Delta$\cons{}. These results suggest that code consistency is not readily predictable from LLMs' internal representations, limiting LLMs' ability to improve explanation consistency through internal representations alone.

\begin{figure}[t]
\begin{center}
\includegraphics[width=\linewidth]{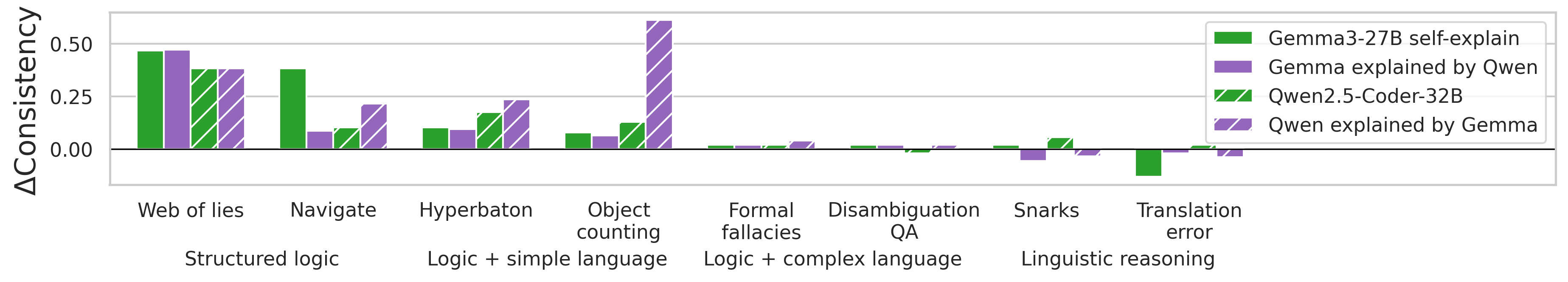}
\end{center}
\caption{$\Delta$ \cons{} of self-explanations and cross-model explanations, where Gemma and Qwen explain their own behavior or each other's, respectively. In 12 of the 16 cases, the two approaches achieve consistency within 0.08 of each other, showing that consistency feedback can benefit both self-explanations and post-hoc explanations.}
\label{fig:cross-model}
\end{figure}

Privileged self-access may not be necessary to obtain consistent explanations. To test this, \Cref{fig:cross-model} compares the code consistency of Gemma and Qwen when explaining their own versus each other's outputs, using one random seed. In 12 of 16 model-task settings, self-explanations and cross-model explanations achieve consistency within 0.08 of each other. In 3 of the remaining 4 settings, cross-model explanations achieve higher consistency. These results suggest that consistency feedback can identify post-hoc explanations that best match a target model's behavior, even when generated by a different LLM.

\begin{figure}[t]
\begin{center}
\includegraphics[width=\linewidth]{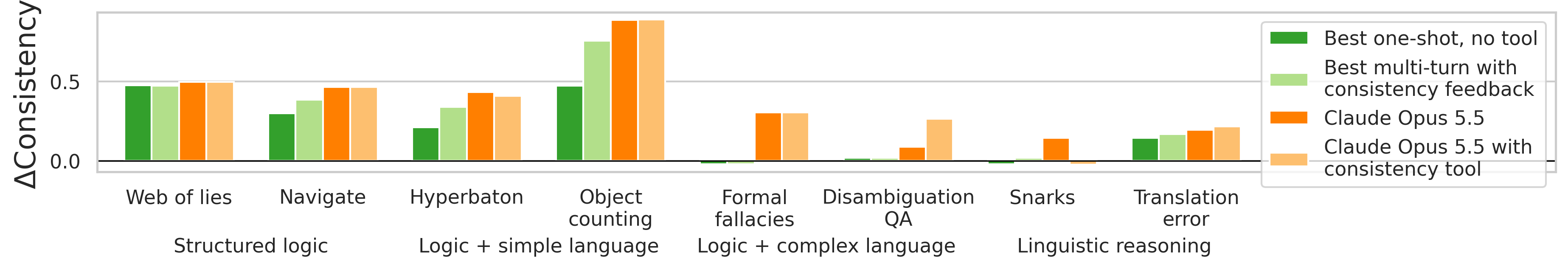}
\end{center}
\caption{$\Delta$ \cons{} across four approaches: (1) one-shot generation, (2) 10-turn iterative refinement with consistency feedback, (3) Claude with code execution, and (4) Claude with execution and consistency feedback. Multi-turn refinement and agentic execution with Claude Opus 5.5 improve consistency over one-shot generation, though consistency remains substantially lower on the last four, more complex tasks.}
\label{fig:advance}
\end{figure}

\paragraph{Verification feedback.} We next investigate whether execution and consistency feedback can improve code explanation consistency in multi-turn agentic settings. We compare four approaches in \Cref{fig:advance} using one random seed: (1) the best one-shot model among the three previously studied; (2) 10-turn iterative refinement with the same models, with the most consistent program generated so far and its consistency provided after each turn; (3) Claude Opus 5.5 with Python execution; and (4) Claude Opus 5.5 with both Python execution and a consistency-evaluation tool. See \Cref{app:implementation-details} for the full implementation details.

\Cref{fig:advance} shows that multi-turn refinement improves consistency on the first four, simpler tasks.
\Cref{fig:multi} further plots code consistency across 10 refinement turns for each model-task setting. On the first four, simpler tasks, $\Delta$\cons{} approaches the maximum in all but two settings. In contrast, $\Delta$\cons{} remains close to zero across turns on the last four, more complex tasks. These results suggest that multi-turn refinement can improve consistency on simpler tasks, but is still insufficient to overcome the challenges of generating consistent explanations for more complex tasks.

Both Claude variants achieve the highest $\Delta$\cons{} (\Cref{fig:advance}) while producing code of comparable complexity to open-source models (\Cref{fig:code-length-with-claude}), highlighting the benefit of execution feedback, with or without the consistency-evaluation tool. We provide examples of Claude-generated code in \Cref{app:code-examples}. \Cref{fig:claude} further compares Claude's task accuracy and explanation consistency. On the first four tasks, Claude achieves 100\% on both metrics, showing that perfectly consistent explanations are possible, although perfect task accuracy makes it impossible to distinguish consistent explanation from accurate task solving. On the last four tasks, Claude maintains task accuracy above 75\% but achieves $\Delta\cons{} \leq 0.3$. These improvements nevertheless exceed the uninformative majority-answer baseline, showing that execution feedback can improve code consistency across all tasks, although consistency remains substantially lower on more complex tasks.

\section{Conclusion}
\label{sec:discussion}

We propose executable code as an automatically verifiable and compact alternative to instance-wise free-text explanations of LLM behavior. We introduce \cons{}, a metric that measures how often an executable explanation reproduces the model's outputs across inputs, and use it to evaluate and select candidate explanations. Experiments across three code-capable LLMs and eight tasks show that highly consistent code explanations are possible in several settings. Code explanations better capture model behavior than free-text CoT in 9 out of 24 settings, while being 6$\times$ to 100$\times$ shorter.  However, high consistency is not the norm. On tasks with more complex linguistic features, code explanations often perform near random chance and tend to rely on longer hard-coded linguistic patterns. Multi-turn refinement improves consistency on simpler tasks, but yields no improvement on more complex ones. Experiments with Claude Opus 5.5 show that equipping an LLM agent with code execution and consistency feedback improves consistency across all tasks, but even Claude struggles to achieve high consistency on more complex tasks. Together, these findings highlight both the advantages of executable explanations as a verifiable and compact approach to explaining LLM behavior and the challenge of reliably producing consistent explanations, motivating future optimization through execution and consistency feedback.

\paragraph{Limitations.} Our evaluation is limited to Python code explanations and fixed generation hyperparameters, which may affect the expressivity, compactness, and consistency of code explanations. We leave exploring these extensions and systematically optimizing generation hyperparameters for future work. Further, our metric measures behavioral consistency but does not establish whether code explanations faithfully describe the model's internal computations.

\paragraph{Interpreting low consistency on linguistically complex tasks.} The low consistency on linguistically complex tasks in our experiments has two possible interpretations. One is that current LLMs lack the ability to generate code that accurately captures their own behavior on tasks involving complex linguistic features. The other is that some linguistic behaviors are inherently difficult to represent through compact, human-interpretable programs, even with more capable explanation methods. Distinguishing between these possibilities is an important direction for future work, both for improving executable explanations and for understanding the limits of compact, human-interpretable representations of LLM behavior.

\section*{Acknowledgments}

This work was partly supported by the NIH/NHLBI Award R01HL148248, NSF Award 1922658 NRT-HDR: FUTURE Foundations, Translation, and Responsibility for Data Science, NSF CAREER Award 2145542, ONR N00014-23-1-2634, NIH R01CA296388, NSF 2404476, Optum, the Gates Foundation, and Apple. This work was also supported by the Institute of Information \& Communications Technology Planning \& Evaluation (IITP) with a grant funded by the Ministry of Science and ICT (MSIT) of the Republic of Korea in connection with the Global AI Frontier Lab International Collaborative Research Agreement. (No. RS-2024-00469482 \& RS-2024-00509279).

\bibliographystyle{plainnat}
\bibliography{main}

\clearpage
\newpage

\appendix

\section*{Appendix}
\section{Figures}
\label{app:figures}

\begin{figure}[H]
\begin{center}
    \includegraphics[width=\linewidth]{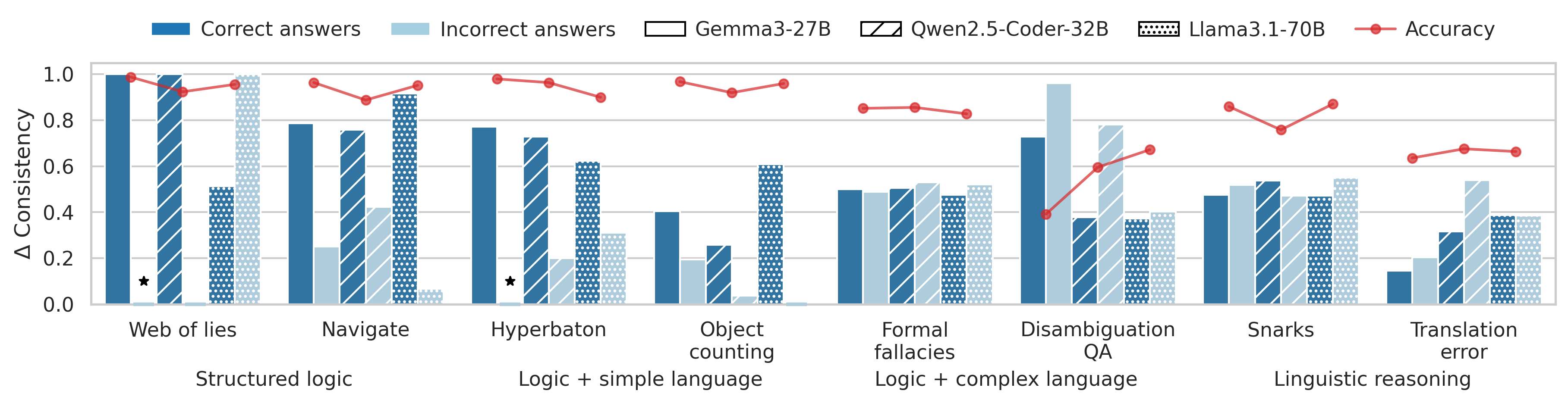}
\end{center}
    \caption{\cons{} stratified by model answer correctness. In 13/24 cases, consistency is no more than $+$0.1 higher for correct than incorrect answers, and in 10/24 cases consistency is higher for incorrect answers, suggesting that code explanations do not systematically explain only the model's correct answer or simply follow the task instruction. In 4 cases, consistency on incorrect answers is 0, but all achieve accuracy $\geq 92\%$, suggesting that imperfectly consistent programs may capture the model's behavior on most examples while failing to capture its occasional mistakes. $\star$ indicate averages computed from fewer than 10 examples, which occur when accuracy is near 100\% and few incorrect answers are available.}
    \label{fig:correct-vs-incorrect}
\end{figure}

\begin{figure}[H]
\begin{center}
\includegraphics[width=0.9\linewidth]{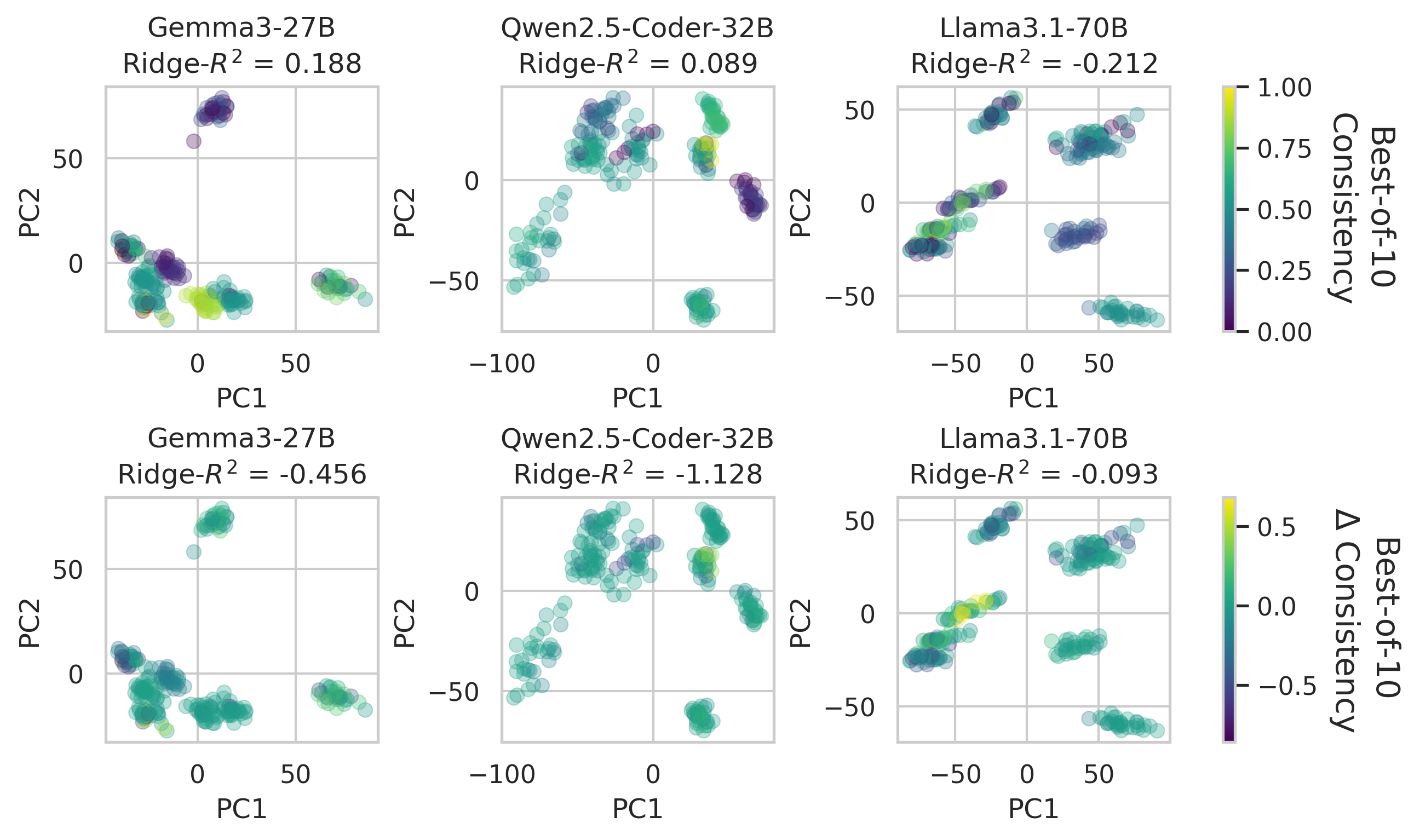}
\end{center}
\caption{PCA visualization of the last hidden state of the final input token across LLMs. Each sample corresponds to one random seed and one set of $N=20$ in-context examples among the $P=10$ code-generation prompts for a model-task setting. The samples are colored by their best-of-$M$ ($M = 10$) code \cons{} (top row) and $\Delta$\cons{} (bottom row). Each panel reports the 5-fold cross-validation $R^2$ for a ridge regression model trained to predict best-of-$10$ $\cons{}$ ($\Delta$\cons{} resp.) from the hidden states. Across all models, ridge regression results in negative $R^2$, and the hidden states show no clear separation by either consistency measure, suggesting that LLM hidden states do not encode information about whether LLMs will produce consistent explanations.}
\label{fig:hidden-state-consistency}
\end{figure}

\begin{figure}[H]
\begin{center}
\includegraphics[width=0.98\linewidth]{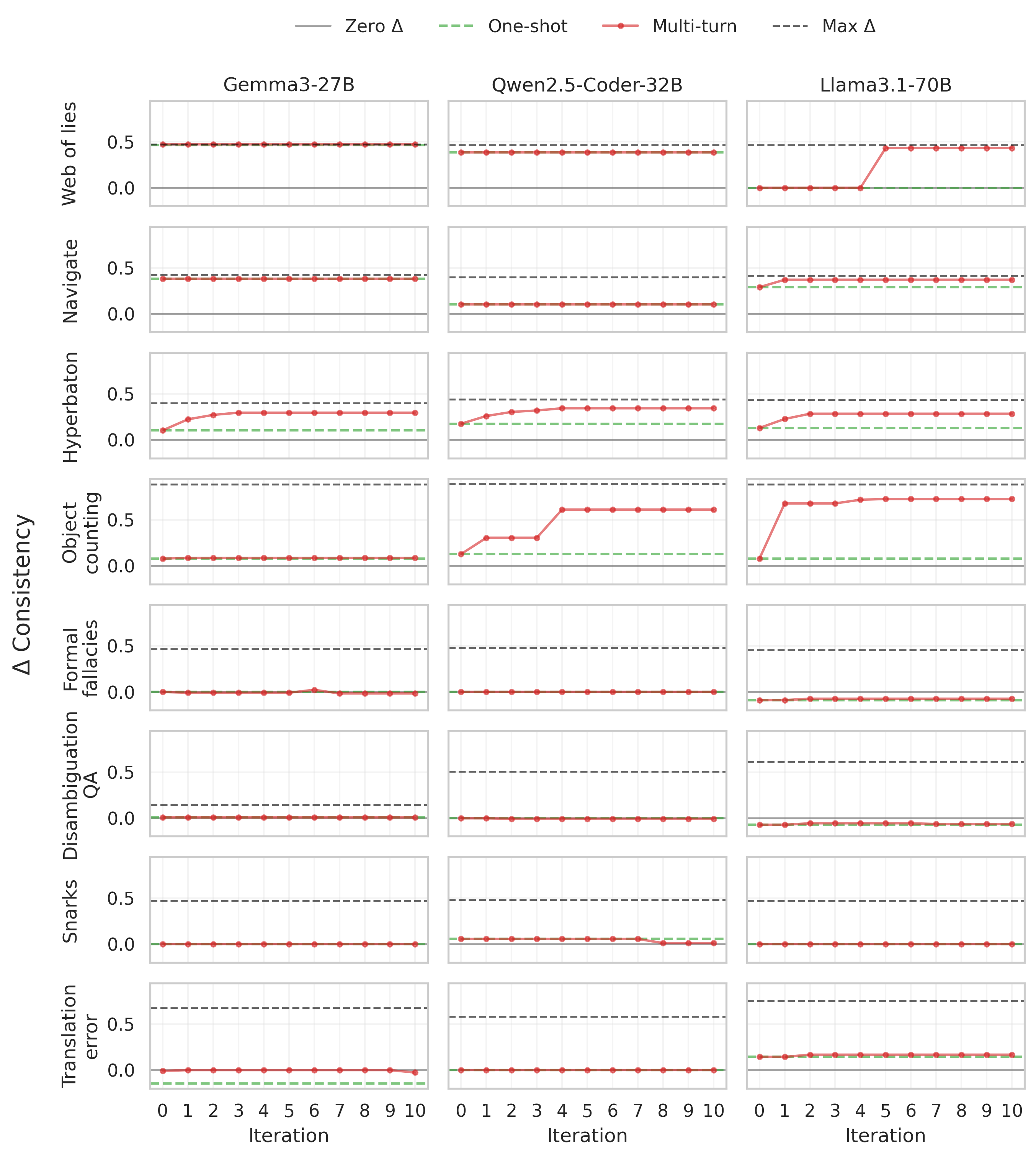}
\end{center}
\caption{Code consistency across 10 refinement turns for each model (column) and task (row) pair. The zero and maximum $\Delta$\cons{} lines are annotated for each setting. On the first four tasks, $\Delta$\cons{} approaches its maximum after 10 turns in all but two settings. On the last four tasks, code consistency remains near its one-shot level, with no improvement across turns.}
\label{fig:multi}
\end{figure}

\begin{figure}[H]
\begin{center}
\includegraphics[width=\linewidth]{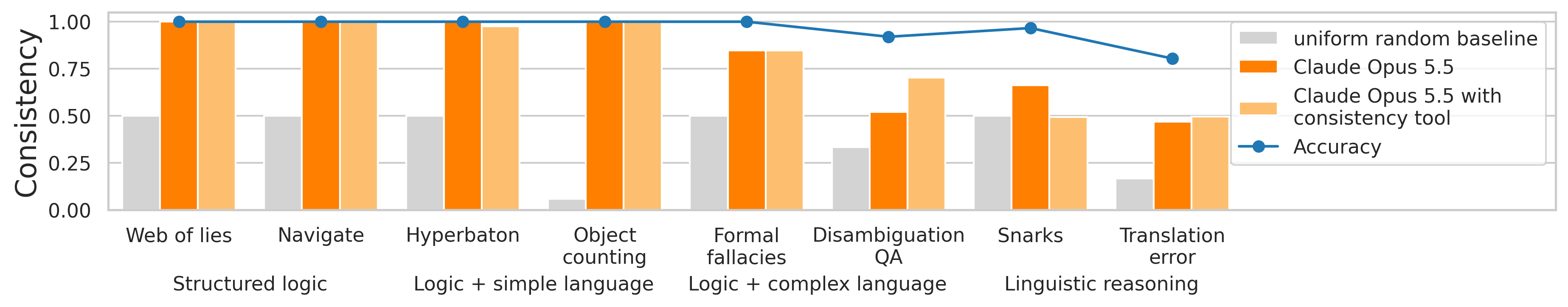}
\end{center}
\caption{Task accuracy and code consistency of Claude, with and without an explicit consistency-evaluation tool, across eight tasks. Both variants achieve close to 100\% on both metrics for the first four tasks. On the last four, accuracy remains above 75\%, but code consistency is at most $+$0.3 higher than the uniform-random baseline.}
\label{fig:claude}
\end{figure}

\begin{figure}[H]
\begin{center}
    \includegraphics[width=\linewidth]{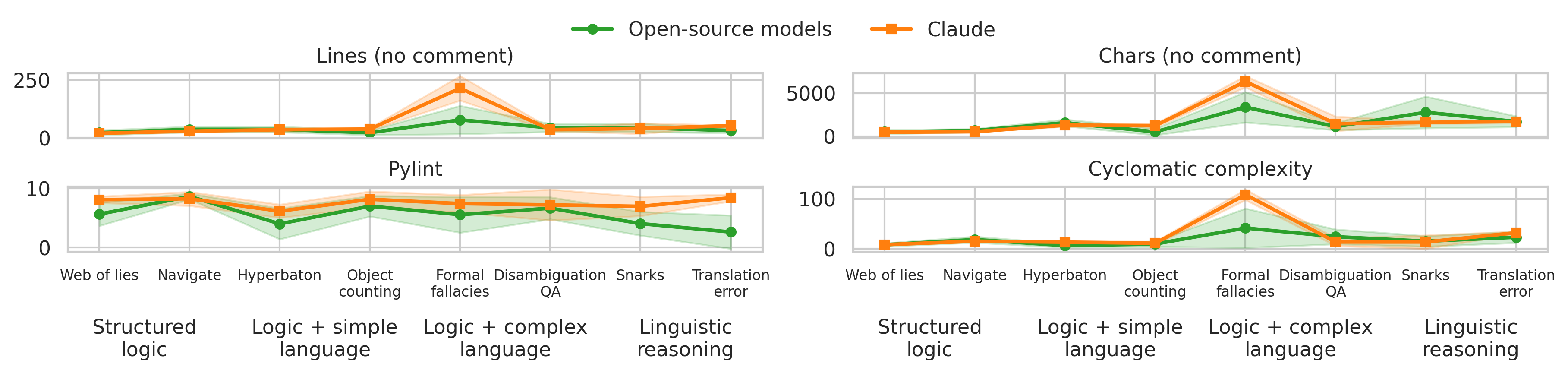}
\end{center}
    \caption{Measures of compactness and interpretability of the best code explanations generated by open-source models and Claude, with standard deviations shown. Across tasks, the complexity of code explanations is similar between open-source models and Claude. Only for the Formal Fallacies task, all models generate explanations with significantly more lines, characters, and higher cyclomatic complexity.}
    \label{fig:code-length-with-claude}
\end{figure}

\section{Implementation details}
\label{app:implementation-details}

\begin{table}[H]
  \caption{Prompt used to generate code explanations for LLM's behavior on a dataset.}
  \label{tab:code-prompt}
  \centering
  \begin{scriptsize}
\begin{tabularx}{\linewidth}{p{1cm}X}
    \toprule
    Role     & Content \\
    \midrule
    System & \texttt{You are given a task along with N example input-output pairs that you previously answered. Your task is to explain your underlying algorithm. The function must correctly handle all tokens that appear in the inputs. Do not ignore any part of the input string.} \\ \midrule
User & \texttt{Task description:\ T; Q: x\_1, A: y\_1; ... ; Q: x\_N, A: y\_N} \\
& \texttt{Write a Python function that reproduces the outputs from the inputs. Think about how to write an algorithm step by step. Make sure that your code does not raise any error when running with the input strings. Do not show example usage. Output the final Python code in the following format:} \\
    \bottomrule
  \end{tabularx}
  \end{scriptsize}
\end{table}

\paragraph{Generate answers.} Model answers are generated with chain-of-thought prompting, temperature 0, repetition penalty 1.0, and a maximum of 2000 new tokens. 

\paragraph{Generate code explanations.} Code explanations are also generated with chain-of-thought prompting, but with temperature 0.1, top-p 0.95, repetition penalty 1.0, and a maximum of 20000 new tokens. We use \Cref{tab:code-prompt} to generate code explanations in Python in the format \texttt{\small solve(x)}, where \texttt{\small x} is the full input text. Although most generated programs follow the \texttt{\small solve(x)} interface, a small number of cases (Translation Error and Disambiguation QA) produce two-argument functions (e.g., \texttt{\small solve(source, translation)}); we handle these edge cases with an input processing step to enable evaluation. For task-instruction-only experiments, we remove the model's answers from the prompt and select code explanations by their consistency with ground-truth answers on the training set. 

\paragraph{Measure compactness and interpretability.} To measure compactness and interpretability of selected code explanations, we measure the number of lines after removing empty lines and comment-only lines, and the number of characters after removing all comments. We compute the Pylint score using the \texttt{\small pylint} library \citep{pylint}, which ranges from 0 to 10 and measures how well Python code conforms to coding standards, including readability, error-proneness, and stylistic conventions. We compute cyclomatic complexity using the \texttt{\small radon} library \citep{radon}, which measures the complexity of a program by computing the number of linearly independent paths through a source code.

\paragraph{Measure free-text explanation sufficiency.} Self-Consistent Sufficiency \citep{nguyen2026sufficiency} evaluates how well an explanation $\be$ explains an LLM's behavior on an input $\bx$ by computing the KL divergence between the model's output distribution given $\bx$ and the average model's output distributions over alternative inputs $\bx'$ that the LLM itself considers possible given $\be$:
\begin{equation}
    \mathcal{S}(\bx, \be, F) := \text{KL} \left[F_{\mby \g \mbx}(\cdot \g \bx) \Big\| \E_{\mbx' \g \mbe} \left[ F_{\mby \g \mbx}(\cdot \g \bx' )\right] \right]
    \label{eq:scsuff}
\end{equation}

We compute free-text sufficiency scores using the same code and hyperparameters as \citet{nguyen2026sufficiency}. Given a code explanation $c$, we evaluate its sufficiency by replacing $\E_{\mbx' \g \mbe} \left[ F_{\mby \g \mbx}(\cdot \g \bx' )\right]$ in \Cref{eq:scsuff} with an output distribution that assigns 99\% probability to $c(\bx)$ and distributes the remaining probability equally among other valid output options. For code explanations, token length is the number of tokens in the generated program. For CoT explanations, token length is the total number of tokens across all CoTs generated for the test set, with one CoT for each input. We evaluated CoT and free-text explanations on the same test split as the main consistency experiments.

\paragraph{Predict \cons{} from hidden states.} For each model-task setting, we extract 80 hidden-state vectors, corresponding to the $P = 10$ code-generation prompts for each of the eight tasks. For each prompt, we compute the best-of-$M$ ($M = 10$) consistency among its generated programs. To predict best-of-10 \cons{} and $\Delta$\cons{} using LLM hidden states, we use standard 5-fold cross-validation with random fold assignment, standardize features using \texttt{StandardScaler}, and fit a ridge regression model with $\alpha=1.0$ using \texttt{\small scikit-learn} library.

\paragraph{Multi-turn refinements.} We use the same initial prompts as in one-shot generation and perform 10 refinement turns. At each turn, we append the best code explanation generated so far and its code consistency score on the training set to the original explanation prompt, and further instruct the model to improve consistency. We evaluate the resulting code explanations on the test set after each turn. Multi-turn refinement experiments are done using one random seed.

\paragraph{Claude agent.} We use Claude Opus 5.5 with access to Python execution and file-editing tools, allowing it to write, execute, test, and revise code explanations. Each agent run is limited to 10 turns and a \$1 budget. In the consistency-feedback variant, we additionally provide a tool that computes code consistency on the training set using the same evaluation function as in our experiments. The agent can use this score to revise its program. We evaluate the final programs on the test set. Claude code experiments are done using one random seed.

\paragraph{Resources.} Computing all metrics on each model-task setting on a single seed takes 2-10 hours on 2-4 A100 GPUs.

\section{Task descriptions and examples}
\label{app:task-examples}

We provide a representative example question from each of the eight BBH \citep{suzgun2023challenging} tasks used in our experiments.

\paragraph{Web of lies.} {\small Question: Sherrie tells the truth. Vernell says Sherrie tells the truth. Alexis says Vernell lies. Michaela says Alexis tells the truth. Elanor says Michaela tells the truth. Does Elanor tell the truth?}

\paragraph{Navigate.} {\small If you follow these instructions, do you return to the starting point? Always face forward. Take 1 step backward. Take 9 steps left. Take 2 steps backward. Take 6 steps forward. Take 4 steps forward. Take 4 steps backward. Take 3 steps right. \\
Options: \\
- Yes \\
- No}

\paragraph{Hyperbaton.} {\small Which sentence has the correct adjective order:\\
Options:\\
(A) midsize old grey Brazilian sweater\\
(B) midsize grey Brazilian old sweater}

\paragraph{Object counting.}
{ \small I have a flute, a piano, a trombone, four stoves, a violin, an accordion, a clarinet, a drum, two lamps, and a trumpet. How many musical instruments do I have?}

\paragraph{Formal fallacies.} {\small ``Here comes a perfectly valid argument: First of all, whoever is a schoolmate of Sondra is not a stepsister of Pricilla. In consequence, whoever is not a stepsister of Pricilla is a schoolmate of Sondra.''\\
Is the argument, given the explicitly stated premises, deductively valid or invalid?\\
Options:\\
- valid\\
- invalid}

\paragraph{Disambiguation QA.} {\small In the following sentences, explain the antecedent of the pronoun (which thing the pronoun refers to), or state that it is ambiguous.\\
Sentence: The patient was referred to the specialist because he had a rare skin condition.\\
Options:\\
(A) The patient had a skin condition\\
(B) The specialist had a skin condition\\
(C) Ambiguous}

\paragraph{Snarks.} {\small Which statement is sarcastic?\\
Options:\\
(A) He's a generous person, trying to promote a charity stream that has raised millions to help kids in need\\
(B) He's a terrible person, trying to promote a charity stream that has raised millions to help kids in need}

\paragraph{Translation error.} {\small The following translations from German to English contain a particular error. That error will be one of the following types: Named Entities: An entity (names, places, locations, etc.) is changed to a different entity. Numerical Values: Numerical values (ordinals or cardinals), dates, and/or units are changed. Modifiers or Adjectives: The modifiers and adjectives pertaining to a noun are changed. Negation or Antonyms: Introduce or remove a negation or change comparatives to their antonyms. Facts: Trivial factual errors not pertaining to the above classes are introduced in the translations. Dropped Content: A significant clause in the translation is removed. Please identify that error. Source: Karl Borromäus Joseph Fürst von Liechtenstein war ein kaiserlicher Feldmarschall.\\
Translation: Charles Borromeo Joseph Prince of Liechtenstein was an judicial field marshal.\\
The translation contains an error pertaining to\\
Options:\\
(A) Modifiers or Adjectives\\
(B) Numerical Values\\
(C) Negation or Antonyms\\
(D) Named Entities\\
(E) Dropped Content\\
(F) Facts}

\section{Code explanation with highest \texorpdfstring{$\Delta$}{Delta}\cons{} for each task}
\label{app:code-examples}

We provide the full code explanation with the highest $\Delta$\cons{} on the test set from the open-source models for each task, alongside the higher-performing of the two Claude variants.

\paragraph{Web of lies.} Gemma3-27B
\begin{scriptsize}
\begin{lstlisting}[language=Python, breaklines=true]
def solve(x):
    """
    This function determines if the last person in a chain of statements tells the truth.
    The chain consists of people making statements about the truthfulness of the previous person.
    The function assumes that exactly one person in the chain is lying.
    """
    statements = x.split(". ")
    n = len(statements)
    
    # Initialize a list to store the truthfulness of each person.
    # True means the person tells the truth, False means they lie.
    truth_values = [True] * n
    
    # Iterate through the statements and update the truth values based on the statements.
    for i in range(n - 1):
        statement = statements[i]
        name = statement.split(" says ")[0]
        
        if "lies" in statement:
            truth_values[i+1] = not truth_values[i]
        else:
            truth_values[i+1] = truth_values[i]
    
    # Check if the last person tells the truth.
    if truth_values[n-1]:
        return "Yes"
    else:
        return "No"
\end{lstlisting}
\end{scriptsize}

Claude Opus 5.5
\begin{scriptsize}
\begin{lstlisting}[language=Python, breaklines=true]
import re
def solve(x):
    s = x.split("Question:")[-1]
    truth = {}
    m = re.search(r"(\w+) (tells the truth|lies)\.", s)
    truth[m.group(1)] = m.group(2) == "tells the truth"
    for a, b, c in re.findall(r"(\w+) says (\w+) (tells the truth|lies)", s):
        truth[a] = truth.get(b, True) == (c == "tells the truth")
    q = re.search(r"Does (\w+) tell the truth", s)
    return "Yes" if truth.get(q.group(1), False) else "No"
\end{lstlisting}
\end{scriptsize}

\paragraph{Navigate.} Gemma3-27B

\begin{scriptsize}
\begin{lstlisting}[language=Python, breaklines=true]
def solve(x):
    instructions = x.lower().split('.')
    x_pos = 0
    y_pos = 0
    direction = 0  # 0: forward, 1: right, 2: backward, 3: left

    for instruction in instructions:
        instruction = instruction.strip()
        if not instruction:
            continue

        words = instruction.split()

        if "take" in words:
            try:
                steps = int(words[words.index("take") + 1])
            except (ValueError, IndexError):
                continue

            if "forward" in words:
                if direction == 0:
                    y_pos += steps
                elif direction == 2:
                    y_pos -= steps
                elif direction == 1:
                    x_pos += steps
                elif direction == 3:
                    x_pos -= steps
            elif "backward" in words:
                if direction == 0:
                    y_pos -= steps
                elif direction == 2:
                    y_pos += steps
                elif direction == 1:
                    x_pos -= steps
                elif direction == 3:
                    x_pos += steps
            elif "right" in words:
                x_pos += steps
            elif "left" in words:
                x_pos -= steps
            else:
                if direction == 0:
                    y_pos += steps
                elif direction == 2:
                    y_pos -= steps
                elif direction == 1:
                    x_pos += steps
                elif direction == 3:
                    x_pos -= steps

        elif "turn" in words:
            if "right" in words:
                direction = (direction + 1) % 4
            elif "left" in words:
                direction = (direction - 1) % 4
            elif "around" in words:
                direction = (direction + 2) % 4

        elif "always face forward" in words:
            pass

    if x_pos == 0 and y_pos == 0:
        return "Yes"
    else:
        return "No"
\end{lstlisting}
\end{scriptsize}

Claude Opus 5.5
\begin{scriptsize}
\begin{lstlisting}[language=Python, breaklines=true]
import re
def solve(x):
    s = x.split('?', 1)[1] if '?' in x else x
    s = s.split('Options:')[0]
    x0 = y0 = 0
    dx, dy = 0, 1
    for ins in [t.strip() for t in s.split('.') if t.strip()]:
        low = ins.lower()
        if low.startswith('turn'):
            if 'around' in low: dx, dy = -dx, -dy
            elif 'left' in low: dx, dy = -dy, dx
            elif 'right' in low: dx, dy = dy, -dx
        elif low.startswith('take'):
            m = re.search(r'\d+', low)
            n = int(m.group()) if m else 0
            if 'backward' in low: vx, vy = -dx, -dy
            elif 'left' in low: vx, vy = -dy, dx
            elif 'right' in low: vx, vy = dy, -dx
            else: vx, vy = dx, dy
            x0 += vx*n; y0 += vy*n
    return 'Yes' if x0 == 0 and y0 == 0 else 'No'
\end{lstlisting}
\end{scriptsize}

\paragraph{Hyperbaton.} Gemma3-27B

\begin{scriptsize}
\begin{lstlisting}[language=Python, breaklines=true]
def solve(x):
    """
    This function determines the sentence with the correct adjective order.

    The adjective order is generally as follows:
    1. Opinion
    2. Size
    3. Physical Quality (shape, age, color)
    4. Origin
    5. Material
    6. Type
    7. Purpose

    The function compares the adjective order in the two given sentences and returns the letter of the sentence with the correct order.
    """

    adj_order = ["opinion", "size", "physical_quality", "age", "shape", "color", "origin", "material", "type", "purpose"]

    def categorize_adjective(adj):
        adj = adj.lower()
        if adj in ["awful", "repulsive", "obnoxious", "terrible", "good", "nice", "lovely"]:
            return "opinion"
        elif adj in ["huge", "big", "small", "medium-size", "normal-size", "massive"]:
            return "size"
        elif adj in ["circular", "square", "triangular", "prismlike", "rectangular"]:
            return "shape"
        elif adj in ["old", "ancient", "old-fashioned", "brand-new", "archaic"]:
            return "age"
        elif adj in ["red", "yellow", "purple", "green", "tan", "silver", "gold", "orange"]:
            return "color"
        elif adj in ["american", "japanese", "pakistani", "iranian", "filipino", "vietnamese", "indonesian"]:
            return "origin"
        elif adj in ["wool", "fiberglass", "wood", "glass"]:
            return "material"
        elif adj in ["exercise", "drinking", "eating", "walking", "smoking", "hiking"]:
            return "purpose"
        else:
            return "type"  # Default to type if not found

    def get_adjective_categories(sentence):
        words = sentence.split()
        adjectives = []
        for word in words:
            category = categorize_adjective(word)
            if category != "type":
                adjectives.append((word, category))
        return adjectives

    sentence1 = x.split("(A) ")[1].split("\n")[0]
    sentence2 = x.split("(B) ")[1].split("\n")[0]

    adjectives1 = get_adjective_categories(sentence1)
    adjectives2 = get_adjective_categories(sentence2)

    def check_order(adjectives):
        order = [adj[1] for adj in adjectives]
        for i in range(len(order) - 1):
            if adj_order.index(order[i]) > adj_order.index(order[i+1]):
                return False
        return True

    if check_order(adjectives1):
        return "A"
    else:
        return "B"
\end{lstlisting}
\end{scriptsize}

Claude Opus 5.5
\begin{scriptsize}
\begin{lstlisting}[language=Python, breaklines=true]
import re

CATS = {
 1: "nice lovely repulsive silly terrible awful ridiculous mysterious obnoxious good bad wonderful beautiful ugly adorable grotesque enormous-looking fantastic horrible",
 2: "little big small tiny massive midsize medium-size large huge enormous giant",
 3: "brand-new new old ancient archaic old-fashioned",
 4: "spherical square rectangular circular triangular prismlike pyramidal",
 5: "purple pink red brown white yellow grey gray tan green blue black orange",
 6: "bangladeshi japanese thai egyptian chinese american indian german russian mexican brazilian turkish ethiopian nigerian filipino vietnamese iranian congolese pakistani indonesian",
 7: "lead steel silver cloth fiberglass wood cardboard rubber iron plastic glass paper leather wool gold",
 8: "exercise smoking snorkeling eating drinking hiking walking",
}
WORDS = {w: c for c, s in CATS.items() for w in s.split()}

def cat(w):
    lw = w.lower()
    if lw in WORDS:
        return WORDS[lw]
    if lw.endswith("ing"):
        return 8
    if w[:1].isupper():
        return 6
    return 1

def ok(phrase):
    words = phrase.split()[:-1]
    cs = [cat(w) for w in words]
    return all(a <= b for a, b in zip(cs, cs[1:]))

def solve(x):
    opts = re.findall(r"\(([A-Z])\)\s*([^\n]+)", x)
    for letter, phrase in opts:
        if ok(phrase.strip()):
            return "(" + letter + ")"
    return "(" + opts[0][0] + ")" if opts else "(A)"
\end{lstlisting}
\end{scriptsize}

\paragraph{Object counting.} Llama3.1-70B

\begin{scriptsize}
\begin{lstlisting}[language=Python, breaklines=true]
def solve(x):
    # Split the input string into individual items
    items = x.split(', ')

    # Initialize a counter for the total number of items
    total = 0

    # Iterate over each item
    for item in items:
        # Check if the item has a quantity specified
        if 'three' in item:
            total += 3
        elif 'four' in item:
            total += 4
        elif 'five' in item:
            total += 5
        elif 'two' in item:
            total += 2
        else:
            # If no quantity is specified, assume it's 1
            total += 1

    return total
\end{lstlisting}
\end{scriptsize}

Claude Opus 5.5
\begin{scriptsize}
\begin{lstlisting}[language=Python, breaklines=true]
import re
NUM={'a':1,'an':1,'one':1,'two':2,'three':3,'four':4,'five':5,'six':6,'seven':7,'eight':8,'nine':9,'ten':10}
CATS={
'vegetable':['lettuce','cauliflower','yam','cabbage','broccoli','potato','potatoe','garlic','carrot','celery','onion'],
'fruit':['peach','raspberr','nectarine','plum','banana','strawberr','blackberr','orange','grape','apple'],
'animal':['fish','snake','mouse','mice','bear','rabbit','pig','goat','cow','snail','donkey','chicken','duck','cat','dog','frog'],
'musical instrument':['trombone','clarinet','drum','violin','piano','flute','accordion','trumpet'],
'object':['fridge','chair','microwave','stove','bed','car','couch','toaster','lamp','table','oven'],
}
def solve(x):
    m=re.search(r'I have (.*)\. How many (.+?) do I have',x)
    items,cat=m.group(1),m.group(2).lower()
    key=None
    for k in CATS:
        if cat.startswith(k): key=k
    total=0
    for part in re.split(r',\s*(?:and\s+)?|\s+and\s+',items):
        w=part.strip().split()
        if not w: continue
        n=NUM.get(w[0].lower(),1)
        noun=' '.join(w[1:]).lower()
        if key and any(re.search(r'\b'+s,noun) for s in CATS[key]):
            total+=n
    return str(total)
\end{lstlisting}
\end{scriptsize}

\paragraph{Formal fallacies.} Llama3.1-70B

\begin{scriptsize}
\begin{lstlisting}[language=Python, breaklines=true]
def solve(x):
    premises = x.split('. ')
    conclusion = premises[-1]
    premises = premises[:-1]
    
    conclusion = conclusion.replace('We may conclude that ', '').replace('We may conclude: ', '').replace('It follows that ', '').replace('So, necessarily, ', '').replace('All this entails that ', '').replace('In consequence, ', '')
    
    conclusion = conclusion.replace('necessarily ', '').replace('not ', 'not_').replace(' is ', '_is_').replace(' or ', '_or_').replace(' and ', '_and_').replace(' every ', 'every_').replace(' whoever ', 'whoever_').replace(' being a ', 'being_a_').replace(' an ', 'an_').replace(' a ', 'a_')
    
    premises = [premise.replace('First premise: ', '').replace('Second premise: ', '').replace('Third premise: ', '') for premise in premises]
    premises = [premise.replace('To start with, ', '').replace('To begin with, ', '').replace('First, ', '').replace('Moreover, ', '').replace('Next, ', '') for premise in premises]
    premises = [premise.replace('Now, ', '').replace('Now ', '') for premise in premises]
    
    premises = [premise.replace(' is ', '_is_').replace(' or ', '_or_').replace(' and ', '_and_').replace(' every ', 'every_').replace(' whoever ', 'whoever_').replace(' being a ', 'being_a_').replace(' an ', 'an_').replace(' a ', 'a_') for premise in premises]
    premises = [premise.replace('not ', 'not_') for premise in premises]
    
    conclusion = conclusion.split(' ')
    premises = [premise.split(' ') for premise in premises]
    
    def is_valid(premises, conclusion):
        # Create a dictionary to store the premises
        premises_dict = {}
        for premise in premises:
            if 'not_' in premise[0]:
                premises_dict[premise[1]] = False
            elif premise[0] == 'every':
                premises_dict[premise[2]] = True
            elif premise[0] == 'whoever':
                premises_dict[premise[2]] = True
            elif '_is_' in premise:
                if premise[0] == 'not':
                    premises_dict[premise[2]] = False
                else:
                    premises_dict[premise[0]] = True
        
        # Check the conclusion
        if 'not_' in conclusion[0]:
            return premises_dict.get(conclusion[1], True) == False
        elif conclusion[0] == 'every':
            return premises_dict.get(conclusion[2], False) == True
        elif conclusion[0] == 'whoever':
            return premises_dict.get(conclusion[2], False) == True
        elif '_is_' in conclusion:
            if conclusion[0] == 'not':
                return premises_dict.get(conclusion[2], True) == False
            else:
                return premises_dict.get(conclusion[0], False) == True
    
    def is_valid_2(premises, conclusion):
        # Create a dictionary to store the premises
        premises_dict = {}
        for premise in premises:
            if '_or_' in premise:
                p1, p2 = premise
                p1 = p1.replace('not_', '')
                p2 = p2.replace('not_', '')
                premises_dict[p1] = True
                premises_dict[p2] = True
            elif '_and_' in premise:
                p1, p2 = premise
                p1 = p1.replace('not_', '')
                p2 = p2.replace('not_', '')
                premises_dict[p1] = True
                premises_dict[p2] = True
            elif 'not_' in premise[0]:
                premises_dict[premise[1]] = False
            elif premise[0] == 'every':
                premises_dict[premise[2]] = True
            elif premise[0] == 'whoever':
                premises_dict[premise[2]] = True
            elif '_is_' in premise:
                if premise[0] == 'not':
                    premises_dict[premise[2]] = False
                else:
                    premises_dict[premise[0]] = True
        
        # Check the conclusion
        if 'not_' in conclusion[0]:
            return premises_dict.get(conclusion[1], True) == False
        elif conclusion[0] == 'every':
            return premises_dict.get(conclusion[2], False) == True
        elif conclusion[0] == 'whoever':
            return premises_dict.get(conclusion[2], False) == True
        elif '_is_' in conclusion:
            if conclusion[0] == 'not':
                return premises_dict.get(conclusion[2], True) == False
            else:
                return premises_dict.get(conclusion[0], False) == True
    
    def is_valid_3(premises, conclusion):
        # Create a dictionary to store the premises
        premises_dict = {}
        for premise in premises:
            if '_or_' in premise:
                p1, p2 = premise
                p1 = p1.replace('not_', '')
                p2 = p2.replace('not_', '')
                premises_dict[p1] = True
                premises_dict[p2] = True
            elif '_and_' in premise:
                p1, p2 = premise
                p1 = p1.replace('not_', '')
                p2 = p2.replace('not_', '')
                premises_dict[p1] = True
                premises_dict[p2] = True
            elif 'not_' in premise[0]:
                premises_dict[premise[1]] = False
            elif premise[0] == 'every':
                premises_dict[premise[2]] = True
            elif premise[0] == 'whoever':
                premises_dict[premise[2]] = True
            elif '_is_' in premise:
                if premise[0] == 'not':
                    premises_dict[premise[2]] = False
                else:
                    premises_dict[premise[0]] = True
        
        # Check the conclusion
        if 'not_' in conclusion[0]:
            return premises_dict.get(conclusion[1], True) == False
        elif conclusion[0] == 'every':
            return premises_dict.get(conclusion[2], False) == True
        elif conclusion[0] == 'whoever':
            return premises_dict.get(conclusion[2], False) == True
        elif '_is_' in conclusion:
            if conclusion[0] == 'not':
                return premises_dict.get(conclusion[2], True) == False
            else:
                return premises_dict.get(conclusion[0], False) == True
    
    if is_valid(premises, conclusion) or is_valid_2(premises, conclusion) or is_valid_3(premises, conclusion):
        return 'valid'
    else:
        return 'invalid'
\end{lstlisting}
\end{scriptsize}

Claude Opus 5.5
\begin{scriptsize}
\begin{lstlisting}[language=Python, breaklines=true]
import re, itertools

ART = r'(?:an? |not |neither |both |no )'

def _split(s, sep):
    return [p for p in re.split(sep, s)]

def parse_pred(s):
    s = s.strip().strip('.').strip()
    s = re.sub(r',\s*too$', '', s)
    s = re.sub(r'^(?:also|however) ', '', s)
    s = s.replace('or, otherwise,', 'or').replace('and, in the same time,', 'and')
    s = s.strip()
    parts = re.split(r',? or (?=' + ART + ')', s)
    if len(parts) > 1:
        return ('or', [parse_pred(p) for p in parts])
    if s.startswith('neither '):
        a, b = s[8:].split(' nor ', 1)
        return ('and', [('not', parse_pred(a)), ('not', parse_pred(b))])
    if s.startswith('not both '):
        return ('not', parse_pred('both ' + s[9:]))
    if s.startswith('both '):
        s = s[5:]
    parts = re.split(r',? and (?=' + ART + ')|, (?=' + ART + ')', s)
    if len(parts) > 1:
        return ('and', [parse_pred(p) for p in parts])
    if s.startswith('not '):
        return ('not', parse_pred(s[4:]))
    s = re.sub(r'^(?:an?|the) ', '', s)
    return ('atom', s.lower().strip())

def atoms(f, acc):
    if f[0] == 'atom': acc.add(f[1])
    elif f[0] == 'not': atoms(f[1], acc)
    elif f[0] in ('and', 'or'):
        for g in f[1]: atoms(g, acc)
    elif f[0] == 'imp':
        atoms(f[1], acc); atoms(f[2], acc)

def ev(f, t):
    k = f[0]
    if k == 'atom': return t[f[1]]
    if k == 'not': return not ev(f[1], t)
    if k == 'and': return all(ev(g, t) for g in f[1])
    if k == 'or': return any(ev(g, t) for g in f[1])
    if k == 'imp': return (not ev(f[1], t)) or ev(f[2], t)

def imp(a, b): return ('imp', a, b)
def neg(a): return ('not', a)

QW = r'(?:everyone|everybody|everything|whoever|whatever|anyone|anything|anybody)'

def parse_sent(s):
    s = s.strip().strip('.').strip()
    s = re.sub(r'^(?:necessarily|then|that),?\s+', '', s)
    s = s.replace('or, otherwise,', 'or').replace('and, in the same time,', 'and')
    if s and s[0].isupper() and not re.match(r'^[A-Z][a-z]+ is ', s) or True:
        pass
    low = s[0].lower() + s[1:] if s else s
    m = re.match(r'^it is (?:false|not the case) that (.+?) is (.+)$', low)
    if m:
        return ('ind', m.group(1).strip(), neg(parse_pred(m.group(2))))
    m = re.match(r'^nobody is neither (.+?) nor (.+)$', low)
    if m:
        return ('all', ('or', [parse_pred(m.group(1)), parse_pred(m.group(2))]))
    m = re.match(r'^(?:nobody|no one|nothing) is (.+)$', low)
    if m:
        return ('all', neg(parse_pred(m.group(1))))
    m = re.match(r'^no (.+?) is (.+)$', low)
    if m:
        subj = m.group(1)
        subs = re.split(r' and no ', subj)
        sf = [parse_subj(x) for x in subs]
        S = sf[0] if len(sf) == 1 else ('or', sf)
        return ('all', imp(S, neg(parse_pred(m.group(2)))))
    m = re.match(r'^(?:every|each|any) (.+?) (?:that|who) is (.+?) is (?:also )?(.+)$', low)
    if m:
        return ('all', imp(('and', [parse_pred(m.group(1)), parse_pred(m.group(2))]), parse_pred(m.group(3))))
    m = re.match(r'^(?:every|each|any) (.+?) is (.+)$', low)
    if m:
        return ('all', imp(parse_pred(m.group(1)), parse_pred(m.group(2))))
    m = re.match(r'^' + QW + r' is none of this: (.+), is (.+)$', low)
    if m:
        lst = m.group(1)
        i = lst.rfind(' or ')
        a, b = lst[:i], lst[i + 4:]
        return ('all', imp(('and', [neg(parse_pred(a)), neg(parse_pred(b))]), parse_pred(m.group(2))))
    m = re.match(r'^' + QW + r' (?:who |that )?is (.+?),? is (?:also |however )?(.+)$', low)
    if m:
        return ('all', imp(parse_pred(m.group(1)), parse_pred(m.group(2))))
    m = re.match(r'^(not )?being (.+?) is (sufficient|necessary) for (not )?being (.+)$', low)
    if m:
        a = parse_pred(m.group(2)); b = parse_pred(m.group(5))
        if m.group(1): a = neg(a)
        if m.group(4): b = neg(b)
        if m.group(3) == 'sufficient':
            return ('all', imp(a, b))
        return ('all', imp(b, a))
    m = re.match(r'^if (?:someone|somebody|something|anyone) is (.+?),? then (?:he|she|that person|they|it) (?:is|are) (.+)$', low)
    if m:
        return ('all', imp(parse_pred(m.group(1)), parse_pred(m.group(2))))
    m = re.match(r'^(?:there is|there exists|there are) (?:somebody|someone|something|anybody) (?:who|that) is (.+)$', low)
    if m:
        return ('some', parse_pred(m.group(1)))
    m = re.match(r'^(?:somebody|someone|something) is (.+)$', low)
    if m:
        return ('some', parse_pred(m.group(1)))
    m = re.match(r'^(?:there is|there exists) (?:an? )?(.+?) (?:who|that) is (.+)$', low)
    if m:
        return ('some', ('and', [parse_pred(m.group(1)), parse_pred(m.group(2))]))
    m = re.match(r'^some (.+?) (?:who|that) is (.+?) is (.+)$', low)
    if m:
        return ('some', ('and', [parse_pred(m.group(1)), parse_pred(m.group(2)), parse_pred(m.group(3))]))
    m = re.match(r'^some (.+?) is (.+)$', low)
    if m:
        return ('some', ('and', [parse_pred(m.group(1)), parse_pred(m.group(2))]))
    m = re.match(r'^(.+?) is (.+)$', s)
    if m:
        return ('ind', m.group(1).strip(), parse_pred(m.group(2)))
    return None

def parse_subj(x):
    return parse_pred(x)

CONC = [r'it follows that', r'we may conclude that', r'we may conclude:', r'we may conclude',
        r'therefore,', r'in consequence,', r'all this entails that', r'so, necessarily,',
        r'hence,', r'from this follows:', r'consequently,', r'thus,', r'it is therefore',
        r'which derives from the fact that', r'so,']
PREM = [r'to start with,', r'to begin with,', r'first of all,', r'first premise:', r'second premise:',
        r'third premise:', r'fourth premise:', r'first,', r'second,', r'third,', r'next,', r'plus,',
        r'now,', r'moreover,', r'finally,', r'furthermore,', r'additionally,', r'besides,',
        r'in addition,', r'also,', r'what is more,', r'all in all,', r'lastly,']

def solve(x):
    try:
        return _solve(x)
    except Exception:
        return 'valid'

def _solve(x):
    t = x
    i = t.find('argument')
    j = t.find(':', i)
    body = t[j + 1:]
    k = body.rfind('"')
    if k > 0: body = body[:k]
    body = body.strip()
    lb = body.lower()
    best = None
    for c in CONC:
        for mm in re.finditer(re.escape(c), lb):
            if best is None or mm.start() < best[0] or (mm.start() == best[0] and mm.end() > best[1]):
                best = (mm.start(), mm.end())
            break
    prem_txt = body[:best[0]]; conc_txt = body[best[1]:]
    prem_txt = prem_txt.replace('St. ', 'St# ')
    sents = [p for p in re.split(r'\.\s*', prem_txt) if p.strip()]
    prems = []
    for p in sents:
        p = p.strip().replace('St# ', 'St. ')
        changed = True
        while changed:
            changed = False
            for pm in PREM:
                if p.lower().startswith(pm):
                    p = p[len(pm):].strip(); changed = True
        prems.append(parse_sent(p))
    conc = parse_sent(conc_txt.strip())
    allf = [p for p in prems + [conc] if p]
    A = set()
    for p in allf: atoms(p[-1], A)
    A = sorted(A)
    types = [dict(zip(A, v)) for v in itertools.product([False, True], repeat=len(A))]
    T = [ty for ty in types if all(ev(p[1], ty) for p in prems if p and p[0] == 'all')]
    if conc[0] == 'some':
        T = [ty for ty in T if not ev(conc[1], ty)]
    if not T: return 'valid'
    consts = {}
    for p in prems:
        if p and p[0] == 'ind': consts.setdefault(p[1], []).append(p[2])
    if conc[0] == 'ind': consts.setdefault(conc[1], []).append(neg(conc[2]))
    for p in prems:
        if p and p[0] == 'some':
            if not any(ev(p[1], ty) for ty in T): return 'valid'
    for c, fs in consts.items():
        if not any(all(ev(f, ty) for f in fs) for ty in T): return 'valid'
    if conc[0] == 'all':
        if not any(not ev(conc[1], ty) for ty in T): return 'valid'
    return 'invalid'
\end{lstlisting}
\end{scriptsize}

\paragraph{Disambiguation QA.} Llama3.1-70B

\begin{scriptsize}
\begin{lstlisting}[language=Python, breaklines=true]
def solve(sentence, options):
    # Tokenize the sentence into words
    words = sentence.split()
    
    # Find the pronoun in the sentence
    pronouns = ['he','she', 'they', 'him', 'her', 'them']
    pronoun = None
    for word in words:
        if word.lower() in pronouns:
            pronoun = word
            break
    
    # If no pronoun is found, return Ambiguous
    if pronoun is None:
        return 'C'
    
    # Determine the antecedent of the pronoun
    if pronoun.lower() == 'they':
        # 'they' can refer to a single person or a group, so it's often ambiguous
        if 'and' in sentence:
            return 'C'
        else:
            # If 'they' is used with a single person, it's likely referring to that person
            for i, word in enumerate(words):
                if word.lower() == 'they':
                    if i > 0 and words[i-1].lower() in ['the', 'a', 'an']:
                        return 'A'
                    else:
                        return 'B'
    elif pronoun.lower() in ['he', 'him']:
        # 'he' and 'him' usually refer to a male
        for i, word in enumerate(words):
            if word.lower() == 'he' or word.lower() == 'him':
                if i > 0 and words[i-1].lower() in ['the', 'a', 'an']:
                    return 'A'
                else:
                    return 'B'
    elif pronoun.lower() in ['she', 'her']:
        #'she' and 'her' usually refer to a female
        for i, word in enumerate(words):
            if word.lower() =='she' or word.lower() == 'her':
                if i > 0 and words[i-1].lower() in ['the', 'a', 'an']:
                    return 'A'
                else:
                    return 'B'
    
    # If the pronoun is not found in the sentence, return Ambiguous
    return 'C'
\end{lstlisting}
\end{scriptsize}

Claude Opus 5.5
\begin{scriptsize}
\begin{lstlisting}[language=Python, breaklines=true]
import re

RULES = [
    ('office', 'C'), ('discuss their favorite', 'C'), ('discuss his favorite', 'C'), ('discuss her favorite', 'C'),
    ('shared a story', 'C'), ('share a story', 'C'), ('shares a story', 'C'),
    ('had cancer', 'C'), ('needed more time', 'C'), ('standing in front', 'C'),
    ('sent the letter', 'C'), ('used big words', 'C'), ('use big words', 'C'), ('arrogant', 'C'),
    ('unable to get in contact', 'C'),
    ('future payments', 'B'), ('could help with tax', 'B'), ('gravely ill', 'B'), ('felt ill', 'B'),
    ('sick', 'B'), ('culinary training', 'B'), ('blueprints', 'B'), ('needed one', 'B'),
    ('may be guilty', 'B'), ('was guilty', 'B'), ('got fined', 'B'), ('get fined', 'B'),
    ('reply', 'B'), ('idea', 'B'), ('needed a lab assistant', 'B'), ('need a lab assistant', 'B'),
    ('stellar performance', 'B'), ('wrote a good book', 'B'), ('smelled awful', 'B'),
    ('might fall', 'B'), ('should floss', 'B'), ('needed to get more rest', 'B'),
    ('should avoid', 'B'), ('send the prices', 'B'), ('need any help', 'B'), ('needed any help', 'B'),
    ('could help', 'B'), ('can help', 'B'),
    ('good mood', 'A'), ('better understood', 'A'), ('too late', 'A'),
    ('working on', 'A'), ('told us that', 'A'), ('tells us that', 'A'),
    ('took the day off', 'A'), ('skin condition', 'A'), ('purchased', 'A'),
    ('forgot to bring', 'A'), ('very busy', 'A'), ('grading policy', 'A'), ('like the cake', 'A'),
    ('likes the cake', 'A'), ('understand the case', 'A'), ('borrow some money', 'A'),
    ('were fired', 'A'), ('was fired', 'A'), ('got fired', 'A'), ('upheld the peace', 'A'),
    ('more experience', 'A'), ('microscope', 'A'), ('filing his taxes', 'A'), ('filing her taxes', 'A'),
    ('filing their taxes', 'A'), ('like to teach', 'A'), ('likes to teach', 'A'),
    ('did not like the design', 'A'), ("didn't like the design", 'A'), ('repairing', 'A'),
    ('history of', 'A'), ('focus on code', 'A'), ('focuses on code', 'A'), ('anything they could help', 'A'),
    ('their secretary', 'A'), ('his secretary', 'A'), ('her secretary', 'A'),
]

def solve(x):
    m = re.search(r'Sentence:\s*(.*?)\n', x)
    s = (m.group(1) if m else x).lower()
    for k, v in RULES:
        if re.search(r'\b' + re.escape(k) + r'\b', s):
            return '(%s)' % v
    if re.search(r'\b(their|his|her)\b', s) and not re.search(r'\b(him|them)\b', s):
        return '(C)'
    if re.search(r'\b(him|her|them)\b', s):
        return '(B)'
    if re.search(r', (and|but) (he|she|they)\b', s):
        return '(C)'
    return '(A)'
\end{lstlisting}
\end{scriptsize}

\paragraph{Snarks.} Qwen2.5-Coder-32B

\begin{scriptsize}
\begin{lstlisting}[language=Python, breaklines=true]
def solve(x):
    # Split the input into lines to separate the question and options
    lines = x.strip().split('\n')
    # The question is the first line
    question = lines[0]
    # The options start from the third line and go until the end
    options = lines[2:]
    # Initialize a variable to keep track of the sarcastic option
    sarcastic_option = None
    
    # Iterate over the options to find the sarcastic one
    for option in options:
        # Split the option into label and statement
        label, statement = option.split(') ', 1)
        # Check for sarcastic indicators in the statement
        if any(indicator in statement for indicator in ['ARE U INSANE', 'Have my down vote', 'naive', 'terrible support', 'pragmatic', 'more expensive', 'less fun', 'liability', 'enjoyable', 'best', 'reddit', 'fun', 'sexist', 'win you points', 'lose you points', 'True', 'rational', 'threaten', 'pleaded', 'highest', 'lowest', 'rich', 'loans', 'suspicious', 'legit', 'perfect']):
            # If the statement contains a sarcastic indicator, mark it as the sarcastic option
            sarcastic_option = label
            break
    
    # Return the label of the sarcastic option in the required format
    return f"<FINAL_ANSWER>\n{label}\n</FINAL_ANSWER>"
\end{lstlisting}
\end{scriptsize}

Claude Opus 5.5
\begin{scriptsize}
\begin{lstlisting}[language=Python, breaklines=true]
import re, difflib

SARC = ["megaconglomerate", "no", "years", "down vote", "person", "worthless", "asus's",
        "more", "generosity", "ever", "usually", "going up", "pleased", "never seen",
        "concrete", "devastating", "10000", "essential", "you don't", "?",
        "obviously", "great", "love", "totally", "sure", "always", "everyone", "best",
        "brilliant", "genius", "perfect", "wonderful", "fun"]
SINCERE = ["exorbitant", "less", "rarely", "displeased", "insufficient", "detrimental",
           "valuable", "liability", "several", "days", "up vote", "startup", "my",
           "self-centeredness", "going down", "i've seen", "stained-glass", "2"]

def _opts(x):
    m = re.search(r"\(A\)\s*(.*?)\s*\(B\)\s*(.*)", x, re.S)
    if not m:
        return "", ""
    return m.group(1).strip(), m.group(2).strip()

def _tok(s):
    return re.findall(r"[\w'\-]+|[^\w\s]", s.lower())

def _diff_text(a, b):
    ta, tb = _tok(a), _tok(b)
    sm = difflib.SequenceMatcher(None, ta, tb, autojunk=False)
    da, db = [], []
    for op, i1, i2, j1, j2 in sm.get_opcodes():
        if op == "equal":
            continue
        da.append(" ".join(ta[max(0, i1-1):i2+1]))
        db.append(" ".join(tb[max(0, j1-1):j2+1]))
    return " | ".join(da), " | ".join(db), i2 - i1 if False else None

def _score(d, core):
    s = 0
    words = " " + d + " "
    for k in SARC:
        if re.search(r"(?<![\w'\-])" + re.escape(k) + r"(?![\w'\-])", d):
            s += 2 if (" " in k or k in core) else 1
    for k in SINCERE:
        if re.search(r"(?<![\w'\-])" + re.escape(k) + r"(?![\w'\-])", d):
            s -= 2 if (" " in k or k in core) else 1
    return s

def solve(x):
    try:
        a, b = _opts(x)
        ta, tb = _tok(a), _tok(b)
        sm = difflib.SequenceMatcher(None, ta, tb, autojunk=False)
        da, db, ca, cb = [], [], [], []
        for op, i1, i2, j1, j2 in sm.get_opcodes():
            if op == "equal":
                continue
            da.append(" ".join(ta[max(0, i1-1):i2+1]))
            db.append(" ".join(tb[max(0, j1-1):j2+1]))
            ca += ta[i1:i2]; cb += tb[j1:j2]
        sa = _score(" | ".join(da), ca)
        sb = _score(" | ".join(db), cb)
        if sa > sb:
            return "(A)"
        if sb > sa:
            return "(B)"
        # fallback: shorter (fewer qualifiers) option tends to be sarcastic
        return "(A)" if len(a) <= len(b) else "(B)"
    except Exception:
        return "(A)"
\end{lstlisting}
\end{scriptsize}

\paragraph{Translation error.} Llama3.1-70B

\begin{scriptsize}
\begin{lstlisting}[language=Python, breaklines=true]
def solve(source, translation):
    # Check for Named Entities error
    if any(word in translation.lower() and word not in source.lower() for word in ['russia', 'new-york', 'philly', 'eduard', 'kumbh', 'rockets', 'germany']):
        return 'D'
    
    # Check for Numerical Values error
    if any(char.isdigit() and source.count(char)!= translation.count(char) for char in source):
        return 'B'
    
    # Check for Modifiers or Adjectives error
    if any(word in translation.lower() and word not in source.lower() for word in ['non', 'commercial','sea', 'gold', 'video']):
        return 'A'
    
    # Check for Negation or Antonyms error
    if 'no' in translation.lower() and 'allen' in source.lower():
        return 'C'
    
    # Check for Facts error
    if'sea' in translation.lower() and 'fluss' in source.lower():
        return 'F'
    
    # Check for Dropped Content error
    if len(source) - len(translation) > 20:
        return 'E'
    
    # If none of the above conditions are met, return 'D' as default
    return 'D'
\end{lstlisting}
\end{scriptsize}

Claude Opus 5.5
\begin{scriptsize}
\begin{lstlisting}[language=Python, breaklines=true]
import re, difflib

def solve(x):
    try:
        src = re.search(r'Source:\s*(.*?)\nTranslation:', x, re.S).group(1).strip()
        tr = re.search(r'Translation:\s*(.*?)\nThe translation contains', x, re.S).group(1).strip()
    except Exception:
        return "(E)"
    sl, tl = src.lower(), tr.lower()
    # Negation / antonyms
    neg_t = re.search(r"\b(not|no|never|n't)\b", tl)
    neg_s = re.search(r"\b(nicht|kein\w*|nie)\b", sl)
    if bool(neg_t) != bool(neg_s):
        return "(C)"
    for w in ['hating', 'wide', 'smallest', 'least', 'worst', 'lowest', 'oldest', 'youngest']:
        if re.search(r'\b' + w + r'\b', tl):
            return "(C)"
    # Numbers
    sn = set(re.findall(r'\d+', src)); tn = set(re.findall(r'\d+', tr))
    if tn - sn:
        return "(B)"
    if sn - tn and not re.search(r'\b(\d+)(st|nd|rd|th)\b', tr):
        return "(E)"
    ords = {'erst': 'first', 'zweit': 'second', 'dritt': 'third', 'viert': 'fourth'}
    for g, e in ords.items():
        if re.search(r'\b' + g, sl) and not re.search(r'\b' + e + r'\b', tl):
            if any(re.search(r'\b' + e2 + r'\b', tl) for e2 in ords.values()) or 'erste bedeutende' in sl:
                return "(B)"
    # Single-letter change
    sl1 = set(re.findall(r'\b([A-Z])\b', src)); tl1 = set(re.findall(r'\b([A-Z])\b', tr))
    if tl1 - sl1:
        return "(F)"
    # Named entities
    stoks = [s for s in re.findall(r'[\w\-]+', src) if s[0].isupper() and len(s) >= 4]
    ttoks = re.findall(r'[\w\-]+', tr)
    for i, t in enumerate(ttoks):
        if not t[0].isupper() or len(t) < 4:
            continue
        tlw = t.lower()
        if tlw in sl:
            continue
        if any(s.lower() in tlw for s in stoks):
            continue
        best = max((difflib.SequenceMatcher(None, tlw, s.lower()).ratio() for s in stoks), default=0)
        if best >= 0.5:
            return "(D)"
    if 'new-york' in tl:
        return "(D)"
    # Parenthetical dropped
    if '(' in src and '(' not in tr:
        return "(A)"
    # Dropped content
    if tr[0].islower():
        return "(E)"
    ratio = len(tr.split()) / max(1, len(src.split()))
    if ratio < 0.9:
        return "(E)"
    return "(F)"
\end{lstlisting}
\end{scriptsize}

\end{document}